\documentclass[journal]{IEEEtran}
\usepackage{amsmath,amsfonts}
\usepackage{algorithmic}
\usepackage{algorithm}
\usepackage{array}
\usepackage[caption=false,font=normalsize,labelfont=sf,textfont=sf]{subfig}
\usepackage{textcomp}
\usepackage{stfloats}
\usepackage{url}
\usepackage{verbatim}
\usepackage{graphicx}
\usepackage{cite}
\PassOptionsToPackage{prologue,dvipsnames,table}{xcolor}
\usepackage[table]{xcolor}
\usepackage{booktabs}
\usepackage[hidelinks]{hyperref}
\usepackage{algorithm}
\usepackage{algorithmic}
\usepackage{colortbl}
\usepackage{svg}
\usepackage{bbding}
\definecolor{mycyan}{cmyk}{.3,0,0,0}
\usepackage{multirow}
\usepackage{todonotes}
\usepackage[switch]{lineno}
\begin{document}

\title{AdaSemiCD: An Adaptive Semi-supervised Change Detection Method Based on Pseudo Label Evaluation}

\author{Lingyan Ran, Dongcheng Wen, Tao Zhuo, Shizhou Zhang, Xiuwei Zhang, Yanning Zhang,~\IEEEmembership{Fellow~IEEE}
\thanks{This work is supported in part by the National Natural Science Foundation of China (62476226), Natural Science Basic Research Program of Shaanxi (2024JC-YBQN-0719), Natural Science Foundation of NingBo (2023J262). \textit{(Corresponding author: Tao Zhuo)}}
\thanks{Lingyan Ran, Dongcheng Wen, Shizhou Zhang, Xiuwei Zhang, and Yanning Zhang are with the School of Computer Science, Shaanxi Provincial Key Laboratory of Speech and Image Information Processing, and the National Engineering Laboratory for Integrated Aerospace-GroundOcean Big Data Application Technology, the Ningbo Institute of Northwestern Polytechnical University, Northwestern Polytechnical University, Xi’an 710072, China. 
Tao Zhuo is in the College of Information Engineering, Northwest A\&F University, Yangling, 712100, China.}
\thanks{Manuscript received April 19, 2021; revised August 16, 2021.}}

\markboth{Journal of \LaTeX\ Class Files,~Vol.~14, No.~8, August~2021}%
{Shell \MakeLowercase{\textit{et al.}}: A Sample Article Using IEEEtran.cls for IEEE Journals}

\IEEEpubid{0000--0000/00\$00.00~\copyright~2021 IEEE}

\maketitle
\begin{abstract}
Change detection(CD) is an essential field in remote sensing, with a primary focus on identifying areas of change in bitemporal image pairs captured at varying intervals of the same region.
The data annotation process for CD tasks is both time-consuming and labor-intensive.
To better utilize the scarce labeled data and abundant unlabeled data, we introduce an adaptive semi-supervised learning method, AdaSemiCD, to improve pseudo-label usage and optimize the training process.
Initially, due to the extreme class imbalance inherent in CD, the model is more inclined to focus on the background class, and it is easy to confuse the boundary of the target object. Considering these two points, we develop a measurable evaluation metric for pseudo-labels that enhances the representation of information entropy by class rebalancing and amplification of ambiguous areas, assigning greater weights to prospective change objects.
Subsequently, to enhance the reliability of samplewise pseudo-labels, we introduce the AdaFusion module, to dynamically identifying the most uncertain region and substituting it with more trustworthy content. 
Lastly, to ensure better training stability, we introduce the AdaEMA module, which updates the teacher model using only batches of trusted samples. 
Experimental results on ten public CD datasets validate the efficacy and generalizability of our proposed adaptive training framework.

\end{abstract}


\section{Introduction}
\IEEEPARstart{C}{hange} detection (CD) has emerged as a significant research focus within the field of remote sensing. 
Its objective is to identify regions of interest that have experienced alterations in bi-temporal image pairs captured at varying times of the same geographical area. 
This method plays a crucial role in remote sensing data analysis and is particularly important in various civilian sectors, such as urban planning \cite{Zhang24Remote,urban2010ISPRS}, rural land management \cite{xing2023Progressive,xing2024Improving}, and disaster assessment \cite{disaster2018earthquake,disaster2018landslide}.


Given that the process of accurately annotating masks for change detection tasks is notably labor-intensive, direct application of traditional supervised learning approaches, such as convolutional neural networks (CNN) \cite{BIFA2024,crossRES2023szw} and Transformers~\cite{bit,changeformer}, to a limited set of labeled data often results in limited performance. 
In response to these challenges, researchers have explored a range of approaches such as weakly supervised change detection (WSCD)~\cite{wu2023fcdgan,WSCD_TGRS24}, unsupervised change detection (USCD)~\cite{USCD_JSTARS23,USCD_TGRS22}, and semi-supervised change detection (SSCD)~\cite{SSL2022ShiZhenWei,SSL2022ShizhenWei2}. 
Although WSCD is cost-efficient, it relies on incomplete or inaccurate labels, which can introduce significant errors and unpredictable noise. USCD, on the other hand, does not require labeled data and leverages the intrinsic patterns present in the data; however, it often faces challenges when tackling specific tasks like classification or detection. Some methods would adopt sample generation strategies~\cite{IAug2022,gen_sample_TGRS21,augmentaion_23,bandara2022ddpm}, which include data augmentation~\cite{augmentaion_23}, generative adversarial networks (GAN)~\cite{gen_sample_TGRS21}, and diffusion models~\cite{bandara2022ddpm}, frequently necessitate the simulation or synthesis of additional data. However, when dealing with limited available samples, these methods may encounter constraints due to insufficient diversity in the generated data, which can diminish the model's ability to generalize. 
As a result, SSCD~\cite{bandaraRCR,peng2021SemiCDNet,zhang2023Semisupervised} emerges as a potentially more effective solution.
\IEEEpubidadjcol
The paradigm of semi-supervised learning (SSL)~\cite{ran2023DTFSeg,ran2024ddf,ran2024semi} aims to enhance CD performance by leveraging the limited available labeled data and the large volume of unlabeled samples. 
Typically, researchers generate pseudo-labels for the unlabeled data to act as guidance during training. These pseudo-labels are often temporary predictions with higher probabilities. 
The most widely used approach is the mean-teacher (MT)~\cite{yuan2024dynamically} framework, which employs a teacher model to generate pseudo-labels that serve as guidance for the student model during the training process. 
The teacher model is subsequently updated using the exponential moving average (EMA)~\cite{MT} of the student model. 
The student model benefits from training on a mix of limited labeled data along with ample pseudo-labeled data, enabling it to identify more important features and resulting in marked enhancements in performance.

While these methods produce acceptable outcomes, significant problems persist: the model indiscriminately treats all samples, irrespective of their quality, and the training process lacks flexibility. 
Firstly, it is evident that unlabeled samples may not always function as efficient `teachers'. Models frequently encounter difficulties in generating reliable high-quality pseudo-labels for intricate samples, which in turn introduces extra noise that can mislead the model's training. 
Subsequently, the EMA updating process does not take into account the quality of samples. Given that training batches may be biased or contain noise~\cite{wei2023noise}, dynamically determining the timing of training updates could contribute to the stability of the training process.
These factors underscore the need for a more precise supervisory approach, failing which it could negatively impact the model's training. 

In this study, we introduce an adaptive learning strategy, AdaSemiCD, designed to improve the accuracy of pseudo-labels and streamline the training process. 
Our framework builds upon the traditional semi-supervised training approach, augmented by two innovative functional modules, AdaFusion and AdaEMA.
Initially, AdaFusion is employed to suppress noise at the individual sample level, thereby enhancing the accuracy of pseudo-labels. Contrary to previous methods like Augseg~\cite{AugSeg} or CutMix~\cite{yun2019cutmix} that relied on entirely random fusion regions, our AdaFusion technique proactively identifies the most uncertain region and substitutes them with reliable content from either labeled datasets or unlabeled datasets of superior quality. 
Following this, we dynamically adjust the rate of parameter updates in the teacher-student model via AdaEMA to ensure improved stability. Although the traditional EMA effectively mitigates fluctuations in model parameters, thereby boosting stability, it persists in uniformly updating after each training iteration, neglecting the model's varying learning outcomes across different iterations when handling a range of training samples. If unlabeled samples contain excessive erroneous information, it can misdirect the model's training. Therefore, our AdaEMA introduces an adaptive selection process for model-level parameter updates, allowing the model to fully integrate superior parameters.

The main contributions of this paper are as follows:
\begin{itemize}
\item{We propose an adaptive SSCD framework named AdaSemiCD, which dynamically improves the pseudo-labels as well as adjusts the training procedure with pseudo label quality assessment.}
\item{We propose an AdaFusion strategy to enhance unreliable unlabeled samples. The fusion region and the trusted contents are selectively chosen with the uncertainty map.}
\item{We propose an AdaEMA parameter update strategy, which updates the teacher model with a batch-wise pseudo-labels improving assessment.}
\item{Experimental results on ten publicly available datasets demonstrate the effectiveness of our method.}
\end{itemize}

\section{Related Work}
\subsection{Semi-supervised Learning}
Semi-supervised learning involves applying supervised learning on a limited amount of labeled data while employing unsupervised learning on a vast set of unlabeled data. SSL is typically divided into three strategies: consistent regularization (CR), self-training, and holistic methods, with the latter integrating the first two strategies within an SSL framework.


CR techniques are grounded in the concept of perturbed consistency, which utilizes the coherence between the model's output after varying degrees of perturbed input data as a training constraint. The three typical consistent regularization frameworks consist of the ${\Pi}$-model~\cite{temprol-ensemle}, the Temporal-ensembling model~\cite{temprol-ensemle}, and the mean-teacher model~\cite{MT}. The ${\Pi}$-model's double-branch network shares the weight; the Temporal-ensembling model amalgamates all the outputs in the time series, with each image's pseudo-labels being the EMA of the previously generated results; the MT model carries out this smoothing operation at the model parameters level. This model has found application in subsequent semi-supervised research across various domains, such as Active-Teacher for semi-supervised object detection~\cite{ActiveTeacher}, \cite{UniMatch, mittal2019semi, iMAS} for semi-supervised general semantic segmentation,~\cite{fixmatch} for image classification, and~\cite{transformation_medical,zhangSemiSAMExploringSAM2023} for semi-supervised medical image segmentation.
There has also been explored in the area of perturbation design, with \cite{xie2020unsupervised} and \cite{bandaraRCR} examining the image-level and feature-level perturbation of CR respectively.

In the realm of self-training methods, the authenticity of pseudo-labels is of paramount importance. This has led to extensive research into the effective selection of high-quality pseudo-labels for supervised learning. Feng et al.\cite{feng2021semi} proposed a method that incrementally adds labeled instances, reduces class bias via Synthetic minority over-sampling technique, and adaptively selects the optimal number of instances to enhance classifier performance. \cite{CAC} employs a set probability threshold as a selection standard. ST++ \cite{yang2022st++} has developed a multi-tier self-training structure, where labels of high confidence are used for self-training repeatedly until all unlabeled samples have been utilized. \cite{u2pl} uses a constant entropy value as the filtering limit.  

CR and self-training are commonly employed together rather than in isolation, creating a holistic strategy for semi-supervised learning. This is illustrated in \cite{UniMatch, AugSeg, iMAS, fixmatch} and \cite{yang2022st++}, as partially described in previous sections.

\subsection{Semi-supervised Change Detection}

Since annotating a large number of images for CD is time-consuming, recent methods mainly focus on the SSCD.
In the realm of CR techniques, the incorporation of the mean-teacher model in CD was first introduced by Bousias et al.~\cite{bousias2021evaluation}. However, the initial outcomes did not show considerable potential, as this SSCD method fell short when compared to a benchmark that exclusively used a restricted quantity of labeled data for entirely supervised learning. Despite increasing labeled data, this disparity continues to expand. Using this as a basis, Mao et al.\cite{mao2023semi} implemented minor and major improvements to the inputs of the teacher and student models, respectively. Furthermore, they formulated an extra teacher-virtual adversararial training component to further reduce the harmful effects of the pseudo label noise.

Additionally, other semi-supervised methods employ either a single model or a two-branch model with shared weights. Such as Sun et al.~\cite{sun2022semisanet} introduce a siamese network. They incorporated additional self-training based on pseudo-labels, employing threshold filtering to eliminate low-quality pseudo-labels. The rationale behind this filtering lies in the potential noise introduced by pseudo-labels with low confidence, which could adversely affect SSL training. Hafner et al.~\cite{hafner2022urban} propose a dual-task SSCD framework that combines building segmentation and change detection, two closely related downstream tasks. They devised a novel consistency constraint between the two change detection masks produced by the siamese segmentation network and the CD network. Bandara et al.~\cite{bandaraRCR} explored feature-based perturbations of the regularization term, applying various data perturbations at the feature level to expand the distribution space of consistency constraints. This approach fully leverages the information embedded in unlabeled samples.
In recent work, Zhang et al.~\cite{zhang2023Semisupervised} imposed two constraints of class consistency and feature consistency on unlabeled datasets. By aligning the feature representations of unlabeled samples on varying and invariant classes, the model could learn from a feature space that is closer to the real distribution, this contributed to the renewal of the best performance record at that time.

Some approaches primarily employ generative adversarial networks to produce data samples and understand feature distributions that approximate real labeled data~\cite{peng2021SemiCDNet,graph2019gan,yang2022gan}. 
Although these efforts have shown notable achievements in SSCD, the notoriously erratic nature of GAN training presents difficulties with hyperparameter tuning. 
Furthermore, the occurrence of gradient vanishing is a common challenge during training. 
Moreover, the discriminator's robust ability to differentiate can cause an imbalance between the generator and discriminator's performance within the GAN unless additional training strategies are applied. 
As a result, reaching an ideal balance is demanding, complicating the practical application of this method. Therefore, we continue the semi-supervised framework leveraging consistency and self-training techniques.

\section{METHODOLOGY}
\begin{figure*}[!t]
\centering
\includegraphics[width=1.8\columnwidth]{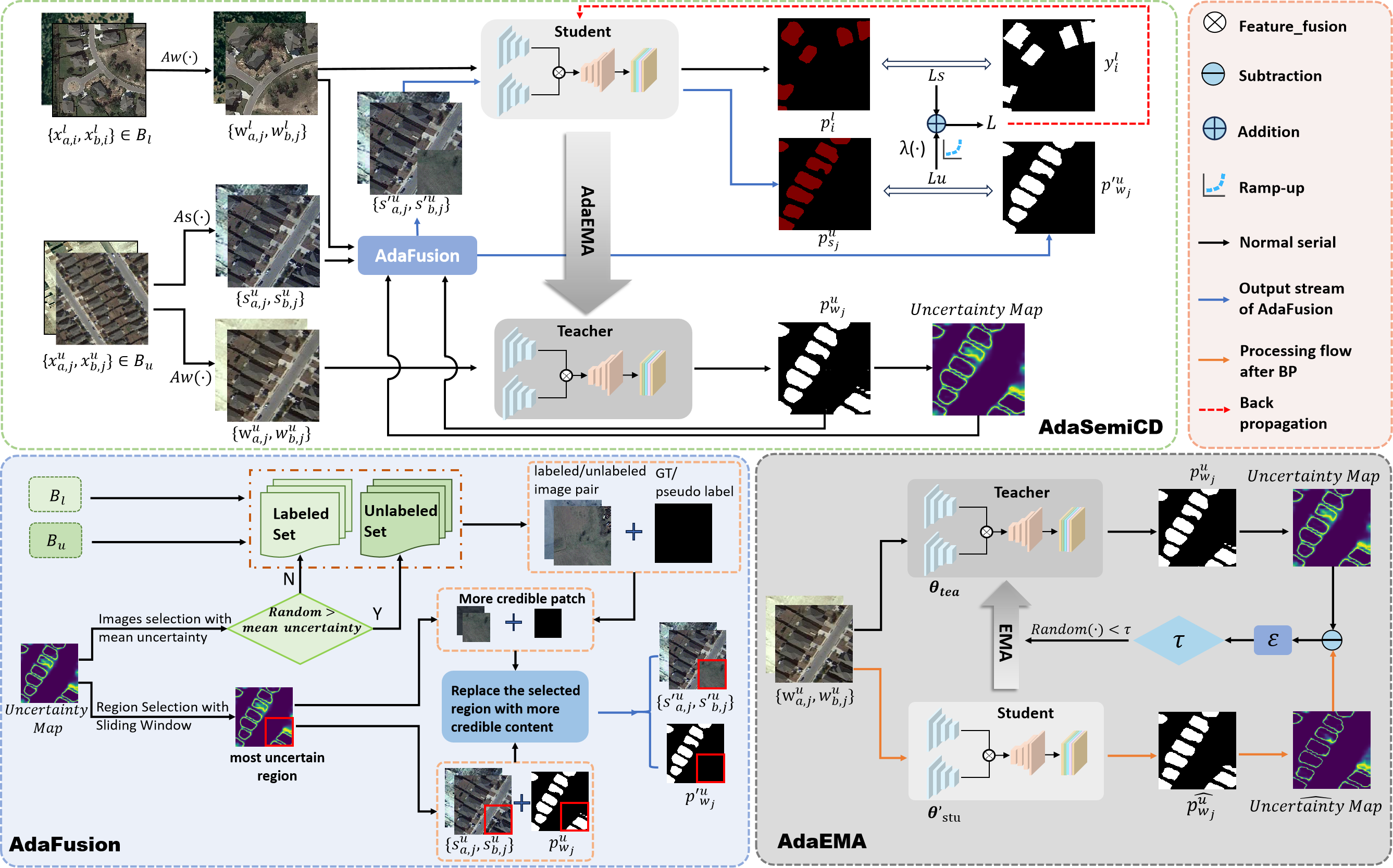}
\caption{The overview of the proposed AdaSemiCD framework and details of two adaptive modules. Our approach is based on the common MT training pipeline for SSCD, and we suggest utilizing the AdaFusion module to produce samples with increased reliability and the AdaEMA module to enhance the efficiency of EMA update.} 
\label{framework}
\end{figure*}

Fig.~\ref{framework} provides a comprehensive summary of our AdaSemiCD framework, aiming to enhance the SSCD performance by leveraging scarce labeled data and a vast quantity of unlabeled samples.
We commence with a general introduction of the framework, followed by an detailed explanation of the uncertainty map used to assess our pseudo-labels, and finally present the specifics of AdaFusion and AdaEMA. 

\subsection{Overview of the AdaSemiCD Framework}\label{overview}
The process of semi-supervised change detection is broadly outlined as follows: We are given a labeled dataset, denoted as $\displaystyle{D}_{l}=\left \{\left \{\smash{x_{a,i}^{l},x_{b,i}^{l}} \right \}, y_{i}^{l}\right \}_{i=1}^{m}$, and an unlabeled dataset \raisebox{0pt}[0pt][0pt]{$\displaystyle{D}_{u}=\left \{x_{a,j}^{u},x_{b,j}^{u} \right \}_{j=1}^{n}$}. Here, $\left \{\left \{\smash{x_{a,i}^{l},x_{b,i}^{l}} \right \}, y_{i}^{l}\right \}$ illustrates the $i$-th pair of labeled images alongside their corresponding true labels, while $\left \{\smash{x_{a,j}^u, x_{b,j}^u}\right \}$ refers to the $j$-th pair of unlabeled images. The subscript $a$ signifies the image from the `Pre'-event period, and $b$ signifies the `Post'-event period. Importantly, the quantity of samples with labels and those without are $m$ and $n$, with $n$ considerably larger than $m$. The aim is for model $M$ to not only derive key insights from ${D}_{l}$ but also to enhance its feature extraction capability using the extensive collection of unlabeled samples in ${D}_{u}$, thus boosting the model's generalization potential.
Ordinarily, samples are subject to either weak augmentation $A_w(\cdot)$ or strong augmentation $A_s(\cdot)$ prior to being passed into the network to guarantee superior generalization performance.

\textbf{Model architecture}: In this study, we employ the widely used MT framework for SSCD tasks. Two integral parts make up the network: a student model, denoted as $M_{stu}$, and a teacher model, represented as $M_{tea}$. Both components possess an identical architecture.
The student model is trained to extract significant features from a small number of labeled samples and a large volume of unlabeled samples, with the aid of optimization via gradient descent methods.
Conversely, the teacher model $M_{tea}$  generates pseudo-labels to guide the student in assimilating unlabeled data, and it is updated using the EMA method.

\textbf{Objectives:} The objective is to minimize the supervised loss $\mathcal{L}_s$ on $D_l$ while ensuring consistency on the disturbed $D_u$ with a minimal $\mathcal{L}_u$.
During training, samples are fed into the network in randomly shuffled batches, $\mathcal{B}_l$ and $\mathcal{B}_u$.

The loss $\mathcal{L}_{s}$ for labeled samples  within a batch $\mathcal{B}_{l}$ is calculated as the cross entropy(CE) of ground truth $y_i^l$ and its prediction $p_i^l$:
\begin{equation}
\label{loss_sup}
\mathcal{L}_{s}=\frac{1}{\left|\mathcal{B}_{l}\right|} \sum_{i=1}^{\left|\mathcal{B}_{l}\right|} \mathrm{CE}\left({p}_{i}^l, y_{i}^l\right),
\end{equation}
where $|\mathcal{B}_{l}|$ represents the mini-batch size, 
${p}_{i}^{l}$ represents the change detection result for the $i$-th pair of images.

The loss $\mathcal{L}_{u}$ for unlabeled samples within a batch $B_u$ is quite similar. Here, we use the pseudo-labels from $M_{tea}$ to supervise the predictions from $M_{stu}$.
The loss $\mathcal{L}_{u}$ is calculated as follows:
\begin{equation}
\label{loss_unsup}
\mathcal{L}_{u}=\frac{1}{\left|\mathcal{B}_{u}\right|} \sum_{j=1}^{\left|\mathcal{B}_{u}\right|} \mathrm{CE}\left(p_{s,j}^u, {p} _{w,j}^u\right),
\end{equation}
where $|\mathcal{B}_{u}|$ represents the size of the unlabeled image mini-batch. ${p}_{w,j}^u = M_{tea}^{\theta}(w_{a,j}^{u},w_{b,j}^{u})$ and ${p}_{s,j}^{u} = M_{stu}^{\theta}(s_{a,j}^{u},s_{b,j}^{u})$ denote the change detection outcomes from the teacher model for the $j$-th pair of weak and strong augmentations of unlabeled images, respectively.

To summarize, the overall loss associated with the AdaSemiCD training process is defined as:
\begin{equation}
 \mathcal{L}=\mathcal{L}_{s}+\lambda(\cdot) \mathcal{L}_{u}.
 \label{totalloss}
\end{equation}
$\lambda$ is the weight of unsupervised loss, which is typically set to a constant value~\cite{fixmatch,transformation_medical}.
For SSCD, we contend that using a constant $\lambda$ could potentially disrupt the training procedure. In the early stages of training, the pseudo-labels for unlabeled data tend to be highly unreliable, and excessive dependence on unsupervised training at this point can inject substantial noise. Conversely, as training progresses into the intermediate and final phases, the model enhances its ability to generate high-quality pseudo labels, diminishing the importance of the limited labeled dataset. This shift warrants a gradual decrease in the emphasis on supervised training compared to unsupervised training over time. To prevent overfitting and refine the feature space, it is crucial to have a systematic approach that dynamically modifies the balance between these two elements in the loss function. We employ a ramp-up approach, defining $\lambda(\cdot)$ as a function that adapts through the course of training, thereby dynamically modulating the balance between supervised and unsupervised training throughout various phases.

\begin{equation}
 \lambda(\cdot)=w_{max}\times e^{-\phi \times\left(1-iter_{cur} / iter_{max}\right)^{2}}
 \label{lambda}
\end{equation}
\begin{equation}
 iter_{max}=\gamma \times iter_total
 \label{iter_max}
\end{equation}
Here, $w_{max}$ represents the maximum weight value of the unsupervised loss, and $\phi$ controls the severity of the ramp-up.
$iter_{cur}$ represents the current iteration cycles; $iter_{max}$ is the total number of ramp-up cycles, calculated by multiplying $\gamma$ (where $0 < \gamma < 1$) by the total number of training iterations, as shown in \ref{iter_max}. After the ramp-up process, the weight of the unsupervised loss stabilizes at $w_{max}$ and no longer changes.
In the early stages of training, this weight is relatively low, so unsupervised training plays a negligible role, but in the middle and later stages of training, this weight gradually increases and the unsupervised loss after weighting exceeds the supervised loss, making unsupervised training the dominant factor.


\textbf{Training strategy:} To reduce the total loss $\mathcal{L}$, the parameters of the student network, denoted as $\theta_{stu}$, are refined via Stochastic Gradient Descent (SGD). Concurrently, the teacher network updates its parameters $\theta_{tea}$ using an exponential moving average calculated from the student network's parameters $\theta_{stu}$ over a time sequence:
\begin{equation}
\theta_{tea} = \beta  \theta_{tea}+(1-\beta )\theta_{stu}
\label{emaequation}
\end{equation}
The hyperparameter $\beta$ acts as a momentum factor, where a higher $\beta$ value leads to a broader moving average window. Generally, $\beta$ is selected to be near 1.0; in this study, for instance, $\beta$ is set at 0.996. 

\textbf{Proposed modules: } The effectiveness of semi-supervised learning depends largely on the quality of pseudo-labels. Nevertheless, it is clear that the previously mentioned procedure does not take into account the varying impact of individual samples on training. This paper concentrates on two critical aspects that are directly related to the generation of pseudo-labels: the initial pair of unlabeled images, and the efficiency of the pseudo-label-generation network, aka the teacher model, in identifying changes. 
To offer more reliable supervisory signals to unlabeled information and reduce training uncertainty, we propose an adaptive training strategy to tackle these two key issues. 
Our initial proposal is to develop a metric that quantifies the uncertainty of pseudo-labels, serving as the basis for adaptive modifications. 
Following that, we recommend applying adaptive modifications at the image level to the unlabeled training samples and suitably integrating reliable contents. 
Additionally, we apply adaptive and selective EMA updates to the teacher network during the training phase to minimize variations, ensuring more consistent and higher-quality pseudo-labels. The details of these methods are discussed in the following sections.

\subsection{Pseudo-label Qualification Metric}
\label{pseudo-label-metric}
To enhance the effectiveness of pseudo-labels by accurately gauging their quality, a crucial step is the implementation of an evaluation metric. This metric plays a key role in identifying reliable labels and determining the value of each training sample.
Unlike labeled image pairs, where true labels serve as a benchmark for assessing the model's performance, pseudo-labels lack such reference points and can only be compared to themselves. To evaluate the quality of pseudo-labels, a common technique is the computation of information entropy~\cite{entropy}, 
\begin{equation}
\label{entropy}
E(x_i)=- P\left(x_{i}\right) \log _{2} P\left(x_{i}\right),
\end{equation}
where $P\left(x_{i}\right)$ is the output probability of a trained model on sample $x_{i}$. 
We propose that reduced information entropy in the predicted values implies greater reliability of the prediction. In contrast, increased information entropy points to a prediction with more uncertainty, indicating a more balanced probability distribution across the pixels and resulting in decreased confidence in the model's predictions.


\begin{figure}[!t]
\includegraphics[width=0.9\columnwidth]{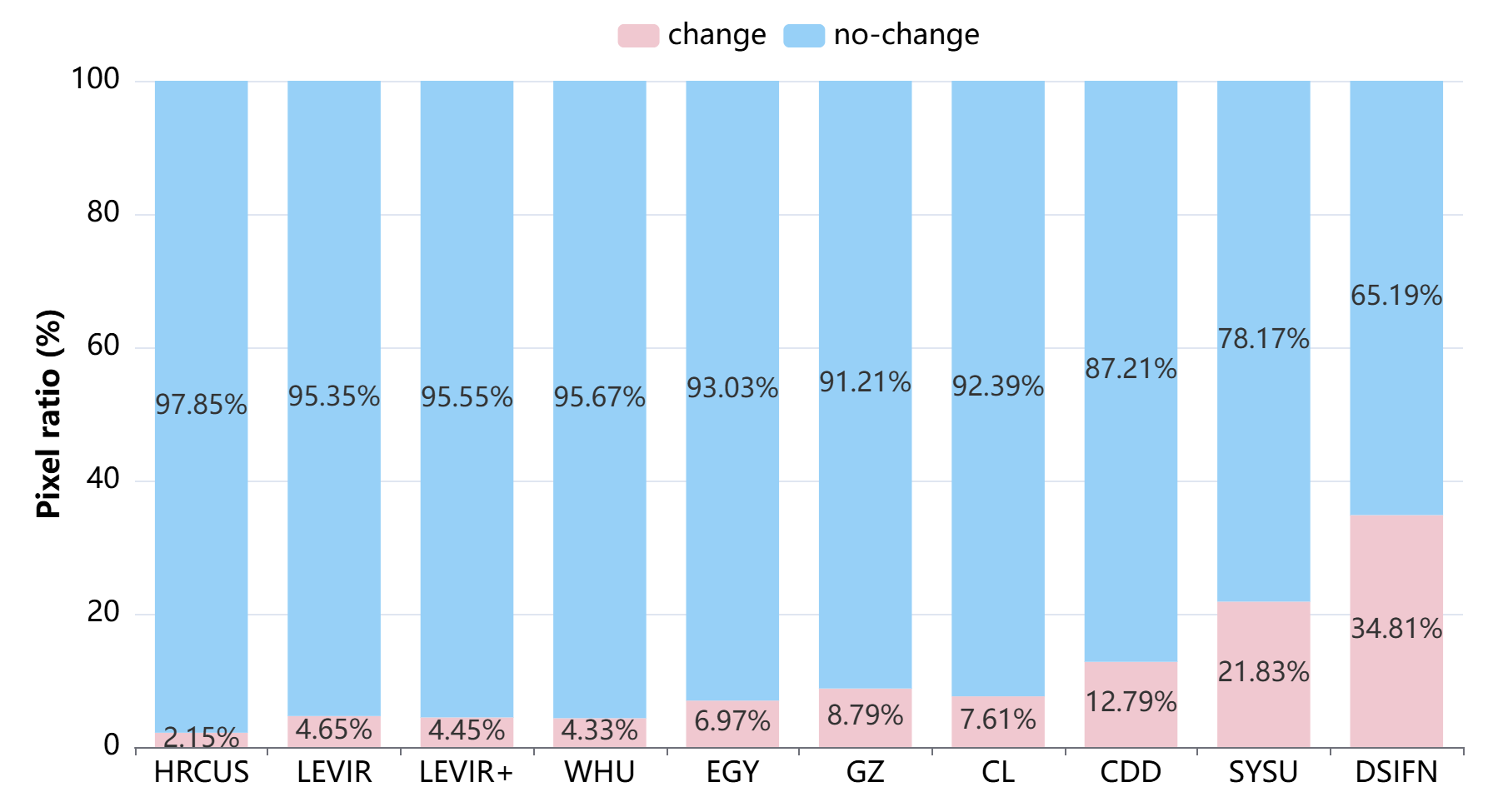}
\caption{Class statistics of common CD datasets. The proportion of the changed/unchanged categories is extremely unbalanced.}
\label{fig_cls}
\end{figure}

In the CD task, directly applying entropy may not yield the optimal results due to the significant challenge posed by class imbalance, as illustrated in Fig.~\ref{fig_cls}. The ratios of changed to unchanged categories are highly skewed. This imbalance may lead our model to learn the target categories inadequately during training while disproportionately capturing feature distributions from the background categories. 
As a result, the model tends to classify pixels as background during inference. 
To reduce the influence of this class imbalance when assessing the quality of pseudo-labels, we assign different weights to the two categories when computing information entropy,
\begin{equation}
\label{weightentropy}
E'\left(X\right)=w_{1}\times E\left(x_i\right)[0]+w_{0}\times E\left(x_i\right)[1],
\end{equation}
where $w_{0}$ and $w_{1}$ represent the proportions of pixels in the current batch that belong to the unchanged and changed categories, respectively. The formulas for their calculation are:
\begin{equation}
\label{w0}
w_{0}=\frac{\sum_{i=1}^{|Bu|} \sum_{k=1}^{H\times W}{(P(x_i)==0)}}{{|Bu|}\times H\times W}
\end{equation}
\begin{equation}
\label{w1}
w_{1}=\frac{\sum_{i=1}^{|Bu|} \sum_{k=1}^{H\times W}{(P(x_i)==1)}}{{|Bu|}\times H\times W}
\end{equation}


Furthermore, pixels from different regions play distinct roles. Pixels that are more crucial, like those found on edges, targets, or areas resembling the background—frequently predicted with higher uncertainty—should be prioritized. To achieve this, we start by enhancing the significance of these key regions by calculating the absolute difference between the prediction probabilities of the two classes,
\begin{equation}
\label{Dx}
D\left(x_i\right)=abs(P\left(x_{i}\right)[1]-P\left(x_{i}\right)[0])
\end{equation}
In this context, $abs$ represents the absolute value function, employed to prevent inconsistencies that could occur due to differing changes before and after a phase. By carrying out a pixel-wise multiplication with the information entropy, we derive the uncertainty map for the image $x_i$,
\begin{equation}
\label{Ix}
U\left(x_i\right) = 1- D\left(x_i\right)\cdot E'\left(x_i\right)
\end{equation}
In areas characterized by low information entropy, this procedure is unlikely to cause significant changes. However, in regions where information entropy is high, the operation enhances the effect.

To summarize, aiming to assess the quality of pseudo-labels in change detection, we propose a quantifiable metric $U$ to evaluate uncertainty. This metric improves overall information entropy by considering elements such as class imbalance and regions of confusion.

\subsection{AdaFusion: Adaptive Sample Fusion}
Image fusion is widely used to augment samples and enhance generalization, with CutMix\cite{yun2019cutmix} and MixUp\cite{zhang2017mixup} being typical examples. 
In this study, our aim is to apply image fusion techniques to exclude unreliable areas from the training samples. This process consists of two steps: region selection, which determines the location for the operation, and image selection,  which specifies the content to be used. 

\textbf{Adaptive selection of fusion region.} 
Contrary to the conventional CutMix technique, which arbitrarily selects blending regions, our approach is more refined. First, we initialize a bounding box of random size, then slide the window to identify the area with the highest total uncertainty using Eq.~\eqref{Ix}, as the region to be fused.
These regions typically include boundaries or complex areas where target identification is difficult. The selection approach guarantees a varied sample and, at the same time, decreases unsupervised noise.

\textbf{Adaptive selection of fusion contents.} 
For the region of maximum uncertainty, we can select a substitute from either the corresponding samples in the labeled set $B_l$ or the unlabeled set $B_u$ with higher reliability. This strategy helps prevent overreliance on labeled samples and further mitigates the risk of overfitting. The selection of fusion content is guided by an adaptive threshold that determines whether to fuse with a labeled image pair. We directly use the computed uncertainty as the threshold, and if the randomly generated probability exceeds the total uncertainty of the sample, labeled images in the training batch are selected as the fusion content. Otherwise, other unlabeled image pairs of higher quality are chosen. It is clear that the higher the reliability of a sample, the better the quality of the pseudo-label. If a sample is considered reliable enough, we avoid fusing it with limited labeled images, which could increase the risk of overfitting. Samples with higher uncertainty contain more noise, and incorporating new content is expected to reduce the noise effectively.

\subsection{AdaEMA: Adaptive EMA Update Strategy}
Within the MT framework, the teacher model's update process involves incorporating the student's exponential moving average over time. 
Our goal is to reach a state of optimal coevolution, where the teacher acts as an aggregate reflection of the student model's progress. 
A pivotal factor in this coevolution is whether the student model advances or declines with each update of the teacher. Evaluating the student model's state is closely tied to the training process's validation phase. 
Typically, after several training epochs, we evaluate the model by using samples from the validation set, performing inference, and comparing the output against true labels to determine accuracy. 
Conducting this validation after every iteration incurs a significant computational cost and extends training duration. One may address this by reducing the validation set size to just a few pairs; however, this risks insufficiently assessing model performance if the sample is too small. The challenge lies in striking a balance between these considerations.

During each training phase, we commence by updating the student model, denoted as $M_{stu}^{\theta}$, according to the strategy outlined in section \ref{overview}, which results in the updated model $M_{stu}^{\theta'}$. Subsequently, we assess both the modified student model, $M_{stu}^{\theta'}$, and the teacher model, $M_{tea}^{\theta}$ on the current set of unlabeled training samples, $\mathcal{B}_u$. Using Eq. \eqref{Ix}, we calculate the corresponding uncertainty maps, referred to as $U_{tea}$ and $U_{stu}$. The changes in the student model's development are captured through fluctuations in uncertainty,
\begin{equation}
\label{evolve}
\varepsilon = \frac{\sum{U_{stu}}-\sum{U_{tea}} }{|B_u|}.
\end{equation}
Additionally, we introduce a probability $\tau$ to regulate model updates according to the reliability of the student model, defined as
\begin{equation}
\label{AdaptEMA}
\tau =\left\{
\begin{array}{cl}
\frac{1}{iter^{2}+\epsilon }&,\quad \varepsilon \leq 0\\
1.0&,\quad \varepsilon > 0 \\
\end{array} \right. ,
\end{equation}
Here, $iter$ denotes the current iteration count and $\epsilon=1e-5$.
When $\varepsilon$ is greater than zero, the student model progresses, allowing a straightforward update of $M_{tea}^{\theta}$. In contrast, if $\varepsilon$ is zero or less, the student model encounters either regression or fluctuation. To gauge the probability of altering $M_{tea}^{\theta}$, we incorporate some randomness.
The adaptive updating details are provided in Algorithm \ref{alg:EMA}.


\begin{algorithm}[tb]
\caption{The AdaEMA algorithm.}
\label{alg:EMA}
\begin{algorithmic}[1] 
\REQUIRE ~~\\ 
    Student model $M_{stu}^{\theta}$, Teacher model $M_{tea}^{\theta}$ \\
    The set of training samples for the current batch, $\mathcal{B} = \left \{\mathcal{B}_l, \mathcal{B}_u\right \}$\\
\ENSURE ~~\\ 
    Updated Teacher model, $M_{tea}^{\theta'}$;
    \newline
    \STATE Calculate the supervised loss $\mathcal{L}_{s}$ on labeled samples $\mathcal{B}_l$ using Eq.~\eqref{loss_sup};\
    \STATE Calculate the unsupervised loss $\mathcal{L}_{u}$ on unlabeled samples $\mathcal{B}_u$ using Eq.~\eqref{loss_unsup};\
    \STATE Update the student model $M_{stu}^{\theta}$ to $M_{stu}^{\theta'}$ using SGD to minimize the total loss, as described in Eq.~\eqref{totalloss}.\
    \STATE Calculate the uncertainty ${U}_{tea}$ of the pseudo-labels generated on $\mathcal{B}_u$ by the teacher model $M_{tea}^{\theta}$ as described in Eq.~\eqref{Ix};\
    \STATE Calculate the uncertainty ${U}_{stu}$ of the pseudo-labels generated on $\mathcal{B}_u$ by updated student model $M_{stu}^{\theta'}$ as described in Eq.~\eqref{Ix};\
    \STATE Calculate the upper bound of the update probability $\tau$ according to Eq.~\eqref{evolve} and Eq.~\eqref{AdaptEMA};
    \IF{$random(\cdot)<\tau$}
        \STATE Update the teacher model to $M_{tea}^{\theta'}$ by EMA;\
    \ELSE
        \STATE $M_{tea}^{\theta'}= M_{tea}^{\theta}$;\
    \ENDIF
\RETURN $M_{tea}^{\theta'}$. 
\end{algorithmic}
\end{algorithm}

\section{Experiment}


\subsection{Experimental Setup}
\subsubsection{Datasets}

\begin{figure*}[!t]
\centering
\includegraphics[width=0.85\textwidth]{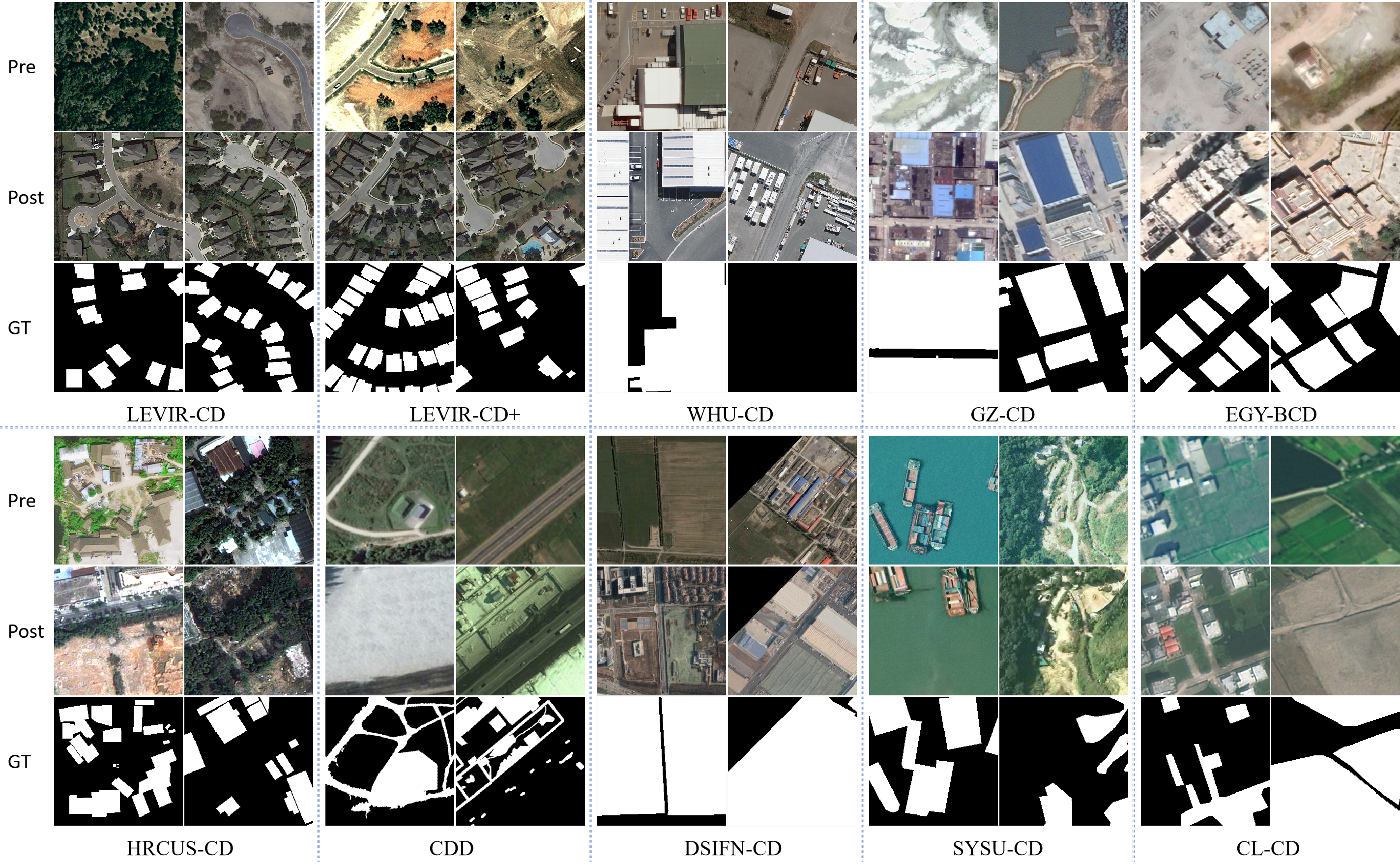}
\caption{Typical examples of ten change detection datasets we use, with Pre-event image, Post-event image, and Ground Truth.}
\label{dataset-samples}
\end{figure*}


Our method is empirically tested on ten benchmark datasets, namely the LEVIR-CD \cite{chen2020levircd}, LEVIR-CD+ \cite{chen2020levircd}, WHU-CD\cite{ji2018whu}, EGY-CD\cite{EGY}, HRCUS-CD\cite{HRCUS}, CDD~\cite{CDD}, GZ-CD\cite{peng2021SemiCDNet}, DSIFN-CD\cite{zhang2020dsifn}, SYSU-CD\cite{SYSU}, and CL-CD\cite{CLCD}.  
As summarized in Table \ref{datasets}, these datasets cover different resolutions (0.03m-2.0m), different data sizes (1-20000 pairs), different annotation categories (binary or multiclass), and different time spans (2-16 years).
Fig.~\ref{dataset-samples}  shows some classic sample images for each dataset.

\begin{table*}[!htbp]
\centering
\caption{Ten publicly available change detection datasets used in the experiment.}
\begin{tabular}{c|c|c|c|c|c|c}
\toprule[1pt]
\rowcolor[HTML]{ECF4FF}
\textbf{Category} &
  \textbf{Datesets} &
  \textbf{Spatial Resolution} &
  \textbf{Size} &
  \textbf{Annotated Samples} &
  \textbf{Time Spans} &
  \textbf{Download} 
  \\ 
  \midrule
 & LEVIR-CD  \cite{chen2020levircd}  & 0.5m & 1024 $\times$ 1024   & 637 & 5 to 14 years &\href{https://justchenhao.github.io/LEVIR/}{\textcolor{blue}{Link}} \\ 
 \cline{2-7} 
 & LEVIR-CD+ \cite{chen2020levircd}  & 0.5m & 1024 $\times$ 1024   & 985  & 5 to 14 years
&\href{https://justchenhao.github.io/LEVIR/}{\textcolor{blue}{Link}}\\ 
 \cline{2-7} 
  &WHU-CD \cite{ji2018whu}&
  0.2m &
  15354$\times$32507 &
  1 &
  2012 to 2016
&\href{http://study.rsgis.whu.edu.cn/pages/download/building_dataset.html}{\textcolor{blue}{Link}}\\
\cline{2-7} 
 & GZ-CD \cite{peng2021SemiCDNet}&
  0.55m &
  Varying &
  19 &
  2006 to 2019
&\href{https://github.com/daifeng2016/Change-Detection-Dataset-for-High-Resolution-Satellite-Imagery}{\textcolor{blue}{Link}}\\
\cline{2-7} 
   & EGY-BCD \cite{EGY}&
  0.25m &
  256 $\times$ 256 &
  6091 &
  2015 to 2022
&\href{https://github.com/oshholail/EGY-BCD}{\textcolor{blue}{Link}}\\
\cline{2-7} 
\multirow{-5}{*}{\textbf{\begin{tabular}[c]{@{}c@{}} Binary\end{tabular}}} &
HRCUS-CD \cite{HRCUS}&
  0.5m &
  256 $\times$ 256 &
  11388 &
  Varing
&\href{https://github.com/zjd1836/AERNet}{\textcolor{blue}{Link}}\\
  \midrule
 & CDD\cite{CDD}   & 0.03m-1.0m & 256 $\times$ 256  & 16000    & Varing 
 &\href{https://drive.google.com/uc?id=0B-IG2NONFdciOWY5QkQ3OUgwejQ&export=download}{\textcolor{blue}{Link}} \\ 
 \cline{2-7} 
 & DSIFN-CD \cite{zhang2020dsifn} &Unknown & 512 $\times$ 512   & 3940    & Unknown
&\href{https://github.com/GeoZcx/A-deeply-supervised-image-fusion-network-for-change-detection-in-remote-sensing-images/tree/master/dataset}{\textcolor{blue}{Link}} \\  
 \cline{2-7} 
 & SYSU-CD\cite{SYSU} &0.5m & 256 $\times$ 256 & 20000  & 2007 to 2014     
 &\href{https://github.com/liumency/SYSU-CD}{\textcolor{blue}{Link}} \\ 
 \cline{2-7} 
\multirow{-4}{*}{\textbf{\begin{tabular}[c]{@{}c@{}} Multiclass\end{tabular}}}
 & CL-CD \cite{CLCD} &0.5-2.0m & 512 $\times$ 512 & 600  & 2017 to 2019  
 &\href{https://github.com/liumency/CropLand-CD}{\textcolor{blue}{Link}} \\
\midrule
\end{tabular}
\label{datasets}
\end{table*}

We employ an identical configuration across all datasets, using 5$\%$, 10$\%$, 20$\%$, and 40$\%$ as the proportions of labeled samples. For LEVIR-CD and WHU-CD, we adopted the semi-supervised partitioning as described in \cite{peng2021SemiCDNet,bandaraRCR,zhang2023Semisupervised}. For the remaining eight datasets, we utilized random partitioning.



\subsubsection{Evaluation Metrics}
For easy comparison with the most advanced techniques, we utilized overall accuracy (OA) to assess general performance. Due to the significant imbalance in CD categories and our main focus on the altered area, we applied the intersection over union for the changed category $IoU^c$. The calculation formulas are provided below:
\begin{equation}
\label{iou}
IoU^c=TP/(TP+FP+FN)
\end{equation}
\begin{equation}
\label{oa}
OA=(TP+TN)/(TP+FP+FN+TN)
\end{equation}
Where $\mathrm{TP}$ represents the positive sample correctly predicted (the correct changing pixel), $\mathrm{TN}$ refers to the negative sample correctly predicted (the correct unchanged pixel), and $\mathrm{FP}$ denotes the positive sample wrongly predicted (the unchanged pixel wrongly detected), $\mathrm{FN}$ represents the negative sample wrongly predicted (the pixel missed as the unchanged pixel).
For both metrics, the larger the value, the better the change detection performance of the model. 

\subsubsection{Implementation Details}


We employ ResNet50+PPM as our change detection framework, as referenced in \cite{bandaraRCR} and \cite{zhang2023Semisupervised}. The learning rate starts at 0.01 and decreases linearly to 1e-4, with momentum kept at 0.9. Training of all competing approaches is conducted using the SGD optimizer over 80 epochs. For both labeled and unlabeled data, the mini-batch size is set at 8. Moreover, the augmentations applied are the same as those in FPA~\cite{zhang2023Semisupervised}, which consist of weak augmentations such as random flip, random resizing from 0.5 to 2.0, and random cropping, along with nine robust augmentations~\cite{cubuk2020randaugment}. A pseudo label threshold of 0.95 is adopted for all models in our studies. When calculating loss, $\phi$ is fixed at 5. The entire set of experiments was executed using PyTorch on four NVIDIA GeForce RTX 3090 GPUs.

\subsection{Compare with the SOTA Methods}
To demonstrate the advantages of our proposed method, we have conducted a comparison with a number of leading semi-supervised change detection techniques~\cite{peng2021SemiCDNet,bandaraRCR,zhang2023Semisupervised,mittal2019semi,vu2019advent}.

\subsubsection{Quantitative Results}

Tables \ref{tab:building-CD} and \ref{tab:mutil-CD} display our experimental outcomes for both the Building CD datasets and the multi-class CD datasets. The `Sup. only' results denote supervised training outcomes from a restricted portion of the labeled dataset, whereas `Oracle' represents full supervision results using the entire training dataset. It's notable that our approach achieves SOTA performance in nearly every partition configuration across these datasets. Remarkably, in most scenarios, all semi-supervised CD approaches outperformed the supervised ones within the same partition setting, confirming the efficacy of semi-supervised methods in leveraging numerous unlabeled training samples. Additionally, the superiority of our method highlights the practical importance of our adaptive strategy for learning more effectively from unlabeled data.
\textbf{Building CD Datasets:} 
As presented in Table \ref{tab:building-CD}, the proposed AdaSemiCD framework demonstrates outstanding performance across nearly all building CD datasets. Notably, on LEVIR-CD, LEVIR-CD+, and WHU-CD datasets, AdaSemiCD achieves a significant improvement in $IoU^c$, with gains of 3.1, 1.3, and 1.7 percentage points, respectively. 
On EGY-CD and HRCUS-CD datasets, AdaSemiCD maintains an overall leading position with average improvements of 0.75 and 0.4 points, respectively. 
However, performance on certain configurations is marginally lower than SOTA methods. 
This limitation can be attributed to the unique challenges posed by these datasets: the EGY-CD dataset suffers from overexposure, which often blends buildings with the background, increasing the difficulty of accurate discrimination. Similarly, the HRCUS-CD dataset is affected by vegetation occluding the changing areas of some buildings. These challenges hinder the model's ability to generate reliable pseudo-labels when trained with a limited number of real labels.

It is evident that AdaSemiCD encounters notable challenges on the GZ-CD dataset, underperforming compared to the FPA method across most experimental configurations, except for maintaining a lead in the 5$\%$ labeled datasets. We attribute this to both the dataset's intrinsic characteristics and the inherent limitations of our approach. Firstly, the GZ-CD dataset has the lowest spatial resolution among the datasets evaluated. This limitation, coupled with the prevalence of small buildings, results in objects that are difficult to discern visually at such resolutions, even for human observers. Secondly, the manual annotations of this dataset are notably coarse, likely due to these resolution constraints. For regions with multiple targets, annotators often delineated broad areas that encompass significant portions of the background, as illustrated in Fig. \ref{dataset-samples}. This introduces substantial noise in the labeled data, especially for building change detection tasks. Lastly, the dataset's suburban setting introduces additional challenges. Urban development and expansion over the dataset's temporal span of over five years result in dramatic differences between the two images, beyond the annotated building change areas. These variations in the background—often so significant that it becomes difficult to ascertain whether the images correspond to the same location—introduce strong non-interesting changes that interfere with the model's ability to focus on the relevant change areas. However, unlike AdaSemiCD, FPA and RCR do not rely solely on one-hot hard label.  Instead, they leverage consistency constraints at the feature level, which effectively mitigates this issue to a significant extent.

Furthermore, as indicated by the experimental results, the overall accuracy of all methods on these binary change detection datasets is notably high, primarily due to the dominance of the background class.  OA often correlates closely with $IoU^c$, where even a slight improvement in OA translates into a relatively significant improvement in $IoU^c$.  Our AdaSemiCD consistently enhances OA across nearly all datasets, as it effectively reduces noise and minimizes the influence of false signals, leading to more accurate predictions.

\begin{table}[tb]
    \centering
    \scriptsize
    \caption{The average quantitative metrics of different CD methods on building change detection datasets. The highlighted parts in blue are the best results, and the underlined ones are the second best results.}
    \resizebox{0.48\textwidth}{!}{
    \begin{tabular}{p{10mm}p{17mm}p{3mm}p{4mm}cp{3mm}p{4mm}cp{3mm}p{4mm}cp{3mm}p{4mm}} %
        \toprule
        \multirow{2}{*}{Dataset} & \multirow{2}{*}{\parbox[c]{.2\linewidth}{Method}} & \multicolumn{2}{c}{5\%} & & \multicolumn{2}{c}{10\%} & & \multicolumn{2}{c}{20\%} & & \multicolumn{2}{c}{40\%}\\ 
        \cmidrule{3-4} \cmidrule{6-7} \cmidrule{9-10} \cmidrule{12-13}
        & & {$IoU^c$} & {OA} && {$IoU^c$} & {OA} & & {$IoU^c$} & {OA} &&{$IoU^c$} & {OA}\\
        \midrule
        \multirow{8}{*}{LEVIR-CD}
        & Sup. only   &   61.0 & 97.60 && 66.8 & 98.13 && 72.3 & 98.44 && 74.9 & 98.60 \\ 
        & AdvEnt\cite{vu2019advent}& 66.1 & 98.08 && 72.3 & 98.45 && 74.6 & 98.58 && 75.0 & 98.60 \\ 
        & s4GAN\cite{mittal2019semi}& 64.0 & 97.89 && 67.0 & 98.11 && 73.4 & 98.51 && 75.4 & 98.62 \\
        & SemiCDNet\cite{peng2021SemiCDNet} & 67.6 & 98.17 && 71.5 & 98.42 && 74.3 & 98.58 && 75.5 & 98.63 \\ 
        & RCR\cite{bandaraRCR}& 72.5 & 98.47 && 75.5 & 98.63 && 76.2 & 98.68 && \underline{77.2} & 98.72 \\
        & FPA\cite{zhang2023Semisupervised}& \underline{73.7} & \underline{98.57} && \underline{76.6} & \underline{98.72} && \underline{77.4} & \underline{98.75} && 77.0 & \underline{98.74} \\     
        \rowcolor{mycyan}
        \multirow{-8}{*}{\cellcolor{white}}& \cellcolor{white}AdaSemiCD   &   \textbf{77.7} & \textbf{98.78} && \textbf{79.4} & \textbf{98.87} && \textbf{80.3} & \textbf{98.92} && \textbf{80.6} & \textbf{98.93} \\
        \cline{2-13}
        & Oracle & \multicolumn{11}{c}{$ IoU^c$=\textcolor{red}{\bf 77.9} and OA=\textcolor{red}{\bf 98.77}} \\
        \bottomrule
        \multirow{8}{*}{LEVIR-CD+}
        & Sup. only   &   52.0 & 97.72 && 58.4 & 98.06 && 66.1 & 98.31 && 66.2 & 98.42 \\ 
        & AdvEnt\cite{vu2019advent}& 52.2 & 97.68 && 59.9 & 98.11 && 65.9 & 98.37 && 68.0 & 98.51 \\ 
        & s4GAN\cite{mittal2019semi}& 46.5 & 97.25 && 51.4 & 97.66 && 62.8 & 98.18 && 67.2 & 98.46 \\
        & SemiCDNet\cite{peng2021SemiCDNet} & 52.6 & 97.66 && 60.7 & 98.24 && 64.8 & 98.37 && 66.1 & 98.38 \\ 
        & RCR\cite{bandaraRCR}& \underline{64.9} & 98.25 && \underline{67.5} & \underline{98.45} && 68.5 & 98.52 && 68.4 & 98.51 \\
        & FPA\cite{zhang2023Semisupervised}& 64.6 & \underline{98.30} && 67.3 & 98.40 && \underline{70.3} & \cellcolor{mycyan}\textbf{98.64} && \underline{69.0} & \underline{98.59} \\     
        \rowcolor{mycyan}
        \multirow{-8}{*}{\cellcolor{white}}& \cellcolor{white}AdaSemiCD   &   \textbf{66.7} & \textbf{98.49} && \textbf{68.8} & \textbf{98.51} && \textbf{70.6} & \cellcolor{white}\underline{98.63} && \textbf{70.9} & \textbf{98.64} \\
        \cline{2-13}
        & Oracle & \multicolumn{11}{c}{$ IoU^c$=\textcolor{red}{\bf 70.5} and OA=\textcolor{red}{\bf 98.63}} \\
        \bottomrule
        \multirow{8}{*}{WHU-CD}
        & Sup. only   &   50.0 & 97.48 && 55.7 & 97.53 && 65.4 & 98.20 && 76.1 & 98.94 \\ 
        & AdvEnt\cite{vu2019advent}& 55.1 & 97.90 && 61.6 & 98.11 && 73.8 & 98.80 && 76.6 & 98.94 \\ 
        & s4GAN\cite{mittal2019semi}& 18.3 & 96.69 && 62.6 & 98.15 && 70.8 & 98.60 && 76.4 & 98.96 \\
        & SemiCDNet\cite{peng2021SemiCDNet} & 51.7 & 97.71 && 62.0 & 98.16 && 66.7 & 98.28 && 75.9 & 98.93 \\ 
        & RCR\cite{bandaraRCR}& 65.8 & 98.37 && \underline{68.1} & \underline{98.47} && \cellcolor{mycyan}\textbf{74.8} & \underline{98.84} && \underline{77.2} & \underline{98.96} \\
        & FPA\cite{zhang2023Semisupervised}& \underline{66.3} & \underline{98.45} && 57.4 & 97.69 && 62.5 & 98.48 && 73.1 & 98.69 \\     
        \rowcolor{mycyan}
        \multirow{-8}{*}{\cellcolor{white}}& \cellcolor{white}AdaSemiCD   &   \textbf{67.8} & \textbf{98.62} && \textbf{70.8} & \textbf{98.70} && \cellcolor{white}\underline{74.7} & \textbf{98.86} && \textbf{79.6} & \textbf{99.13} \\
        \cline{2-13}
        & Oracle & \multicolumn{11}{c}{$ IoU^c$=\textcolor{red}{\bf 85.5} and OA=\textcolor{red}{\bf 99.38}} \\
        \bottomrule
        \multirow{8}{*}{GZ-CD}
        & Sup. only   &   47.5 & 93.56 && 51.4 & 94.26 && 58.0 & 95.65 && 66.3 & 96.62 \\ 
        & AdvEnt\cite{vu2019advent}& 48.6 & 94.39 && 50.9 & 94.89 && 60.2 & 95.79 && 66.2 & 96.58 \\ 
        & s4GAN\cite{mittal2019semi}& 50.8 & 94.38 && 52.4 & 94.98 && 60.8 & 95.94 && 64.2 & 96.39 \\
        & SemiCDNet\cite{peng2021SemiCDNet} & 48.4 & 93.58 && 49.7 & 94.79 && 59.0 & 95.66 && 66.3 & 96.57 \\ 
        & RCR\cite{bandaraRCR}& 50.8 & 93.82 && 50.8 & 94.69 && 62.5 & 96.07 && 67.8 & 96.61 \\
        \rowcolor{mycyan}
        \multirow{-7}{*}{\cellcolor{white}}& \cellcolor{white}
        FPA\cite{zhang2023Semisupervised}& \cellcolor{white}51.2 & \cellcolor{white}93.92 && \textbf{58.9} & \textbf{95.78} && \textbf{63.1} & \textbf{96.26} && \textbf{68.2} & \textbf{96.82} \\     
        \multirow{-8}{*}{\cellcolor{white}}& \cellcolor{white}AdaSemiCD   &   \cellcolor{mycyan}\textbf{51.6} & \cellcolor{mycyan}\textbf{94.56} && \underline{57.1} & \underline{95.57} && \underline{62.4} & \underline{96.21} && \underline{68.0} & \underline{96.75} \\
        \cline{2-13}
        & Oracle & \multicolumn{11}{c}{$ IoU^c$=\textcolor{red}{\bf 69.0} and OA=\textcolor{red}{\bf 96.93}} \\
        \bottomrule
        \multirow{8}{*}{EGY-CD}
        & Sup. only   &   49.8 & 95.73 && 54.6 & 96.38 && 61.4 & 96.83 && 65.1 & 97.25 \\ 
        & AdvEnt\cite{vu2019advent}& 52.7 & 96.01 && 57.8 & 96.58 && 62.6 & 96.86 && 64.0 & 97.19 \\ 
        & s4GAN\cite{mittal2019semi}& 52.9 & 95.94 && 58.6 & 96.50 && 64.7 & 97.09 && 64.9 & 97.27 \\
        & SemiCDNet\cite{peng2021SemiCDNet} & 52.4 & 96.00 && 57.9 & 96.31 && 62.8 & 96.95 && 63.8 & 97.19 \\ 
        & RCR\cite{bandaraRCR}& \underline{58.1} & 96.50 && 59.9 & 96.77 && 63.9 & 97.08 && 64.2 & 97.18 \\
 
        \multirow{-7}{*}{\cellcolor{white}}& \cellcolor{white}
        FPA\cite{zhang2023Semisupervised}& 57.5 & \underline{96.52} && \underline{60.1} & \cellcolor{mycyan}\textbf{96.86} &\cellcolor{mycyan}& \cellcolor{mycyan}\textbf{65.2} & \cellcolor{mycyan}\textbf{97.25} && \underline{65.7} & \underline{97.34} \\  
        
        \rowcolor{mycyan}
        \multirow{-8}{*}{\cellcolor{white}}& \cellcolor{white}AdaSemiCD   &   \textbf{59.0} & \textbf{96.55} && \textbf{60.5} & \cellcolor{white}\underline{96.80} & \cellcolor{white} & \cellcolor{white}\underline{65.0} & \cellcolor{white}\underline{97.20} & \cellcolor{white}& \textbf{67.4} & \textbf{97.39} \\
        \cline{2-13}
        & Oracle & \multicolumn{11}{c}{$ IoU^c$=\textcolor{red}{\bf 67.6} and OA=\textcolor{red}{\bf 97.54}} \\
        \bottomrule
        \multirow{8}{*}{HRCUS-CD}
        & Sup. only   &   29.5 & 98.11 && 36.0 & 98.45 && 43.4 & 98.68 && 48.9 & 98.84 \\ 
        & AdvEnt\cite{vu2019advent}& 29.1 & 98.11 && 36.9 & 98.40 && 42.5 & 98.61 && 48.8 & 98.71 \\ 
        & s4GAN\cite{mittal2019semi}& 25.0 & 97.86 && 28.2 & 98.24 && 40.1 & 98.62 && 50.3 & 98.85 \\
        & SemiCDNet\cite{peng2021SemiCDNet} & 28.4 & 98.00 && 34.7 & 98.44 && 44.1 & 98.68 && 48.5 & 98.74 \\ 
        & RCR\cite{bandaraRCR}& \underline{36.1} & 98.36 && 42.1 & \underline{98.69} && 45.3 & 98.76 && 49.6 & 98.66 \\

        & FPA\cite{zhang2023Semisupervised}& 35.2 & \underline{98.37} && \cellcolor{mycyan}\textbf{43.7} & 98.65 && \underline{46.7} & \underline{98.82} && \cellcolor{mycyan}\textbf{51.2} & \textbf{98.81} \\  
        
        \rowcolor{mycyan}
        \multirow{-8}{*}{\cellcolor{white}}& \cellcolor{white}AdaSemiCD   &  \textbf{37.8} & \cellcolor{mycyan}\textbf{98.59} && \cellcolor{white}\underline{42.6} & \textbf{98.70} && \textbf{48.1} & \textbf{98.84} && \cellcolor{white}\underline{50.8} & \underline{98.87} \\
        \cline{2-13}
        & Oracle & \multicolumn{11}{c}{$ IoU^c$=\textcolor{red}{\bf 59.0} and OA=\textcolor{red}{\bf 99.06}} \\
        \bottomrule
    \end{tabular}
    }
    \label{tab:building-CD}
\end{table}

\begin{table}[tb]
    \centering
    \scriptsize
    \caption{The average quantitative metrics of different CD methods on multiclass change detection datasets. The highlighted parts in blue are the best results, and the underlined ones are the second best results.}
    \resizebox{0.48\textwidth}{!}{
    \begin{tabular}{p{10mm}p{17mm}p{3mm}p{4mm}cp{3mm}p{4mm}cp{3mm}p{4mm}cp{3mm}p{4mm}} %
        \toprule
        \multirow{2}{*}{Dataset} & \multirow{2}{*}{\parbox[c]{.2\linewidth}{Method}} & \multicolumn{2}{c}{5\%} & & \multicolumn{2}{c}{10\%} & & \multicolumn{2}{c}{20\%} & & \multicolumn{2}{c}{40\%}\\ 
        \cmidrule{3-4} \cmidrule{6-7} \cmidrule{9-10} \cmidrule{12-13}
        & & {$IoU^c$} & {OA} && {$IoU^c$} & {OA} & & {$IoU^c$} & {OA} &&{$IoU^c$} & {OA}\\
        \midrule
        \multirow{8}{*}{CDD-CD}
        & Sup. only   &   60.4 & 94.25 && 67.9 & 95.46 && 75.6 & 96.59 && 82.3 & 97.56 \\ 
        & AdvEnt\cite{vu2019advent}& 63.3 & 94.65 && 71.2 & 96.01 && 79.3 & 97.14 && \underline{82.9} & \underline{97.66} \\ 
        & s4GAN\cite{mittal2019semi}& 62.3 & 94.69 && 71.0 & 95.94 && 79.0 & 97.10 && 82.8 & 97.63 \\
        & SemiCDNet\cite{peng2021SemiCDNet} & 63.5 & 94.68 && 71.2 & 95.99 && 79.1 & 97.13 && 82.8 & 97.63 \\ 
        & RCR\cite{bandaraRCR}& 67.6 & 95.40 && \underline{75.5} & \underline{96.57} && \underline{80.2} & \underline{97.26} && 82.7 & 97.61 \\
        & FPA\cite{zhang2023Semisupervised}& \underline{68.9} & \underline{95.66} && 74.9 & 96.55 && 79.7 & 97.20 && 81.1 & 97.37 \\     
        \rowcolor{mycyan}
        \multirow{-8}{*}{\cellcolor{white}}& \cellcolor{white}AdaSemiCD   &   \textbf{70.1} & \textbf{95.89} && \textbf{77.3} & \textbf{96.89} && \textbf{82.1} & \textbf{97.56} && \textbf{83.9} & \textbf{97.80} \\
        \cline{2-13}
        & Oracle & \multicolumn{11}{c}{$ IoU^c$=\textcolor{red}{\bf 87.8} and OA=\textcolor{red}{\bf 98.10}} \\
        \bottomrule
        \multirow{8}{*}{DSIFN-CD}
        & Sup. only   &   34.8 & 78.34 && \underline{38.9} & 83.41 && 40.2 & 87.00 && 39.6 & 87.00 \\ 
        & AdvEnt\cite{vu2019advent}& 31.8 & 77.83 && 36.3 & 83.86 && 40.8 & 85.92 && 37.4 & 86.31 \\ 
        & s4GAN\cite{mittal2019semi}& 36.6 & \underline{84.10} && 34.8 & \underline{86.87} && 37.9 & \cellcolor{mycyan}\textbf{87.69} && \underline{40.1} & 86.52 \\
        & SemiCDNet\cite{peng2021SemiCDNet} & 33.6 & 78.60 && 37.9 & 84.18 && 39.1 & 86.77 && 39.1 & \underline{87.05} \\ 
        & RCR\cite{bandaraRCR}& 26.7 & 83.78 && 32.9 & 86.05 && 40.8 & 86.70 && 36.7 & 86.08 \\
        & FPA\cite{zhang2023Semisupervised}& \cellcolor{mycyan}\textbf{39.2} & \cellcolor{mycyan}\textbf{84.27} && 38.5 & \cellcolor{mycyan}\textbf{87.12} && 36.0 & \underline{87.41} && 35.8 & 86.50 \\     
        \multirow{-8}{*}{\cellcolor{white}}& \cellcolor{white}AdaSemiCD   &   \cellcolor{white}\underline{36.9} & \cellcolor{white}80.46 && \cellcolor{mycyan}\textbf{39.2} & 82.94 && \cellcolor{mycyan}\textbf{41.1} & 85.45 && \cellcolor{mycyan}\textbf{45.1} & \cellcolor{mycyan}\textbf{87.12} \\
        \cline{2-13}
        & Oracle & \multicolumn{11}{c}{$ IoU^c$=\textcolor{red}{\bf 58.1} and OA=\textcolor{red}{\bf 90.82}} \\
        \bottomrule
        \multirow{8}{*}{SYSU-CD}
        & Sup. only   &   62.9 & 89.57 && 64.4 & 90.18 && 66.0 & 90.82 && 66.4 & 90.93 \\ 
        & AdvEnt\cite{vu2019advent}& 61.2 & 89.36 && 64.5 & 90.18 && 65.7 & 90.35 && 68.3 & 91.24 \\ 
        & s4GAN\cite{mittal2019semi}& 64.4 & 90.02 && 66.5 & 90.48 && 66.9 & 90.26 && 68.2 & 91.51 \\
        & SemiCDNet\cite{peng2021SemiCDNet} & 61.7 & 89.32 && 64.8 & 90.25 && 66.7 & 90.97 && 67.0 & 91.08 \\ 
        & RCR\cite{bandaraRCR}& 62.5 & 89.76 && 66.0 & 90.75 && 64.1 & 90.22 && 65.3 & 90.56 \\
        & FPA\cite{zhang2023Semisupervised}& \cellcolor{mycyan}\textbf{67.7} & 90.95 && \underline{68.3} & \underline{91.09} && \underline{70.1} & \underline{92.01} && \underline{69.3} & \cellcolor{mycyan}\textbf{91.97} \\     
        \rowcolor{mycyan}
        \multirow{-8}{*}{\cellcolor{white}}& \cellcolor{white}AdaSemiCD   &   \cellcolor{white}\underline{67.5} & \textbf{91.16} && \textbf{68.7} & \textbf{91.59} && \textbf{70.1} & \textbf{92.03} && \textbf{69.9} & \cellcolor{white}\underline{91.90} \\
        \cline{2-13}
        & Oracle & \multicolumn{11}{c}{$ IoU^c$=\textcolor{red}{\bf 68.2} and OA=\textcolor{red}{\bf 91.64}} \\
        \bottomrule
        \multirow{8}{*}{CL-CD}
        & Sup. only   &   18.1 & 91.90 && 31.4 & 92.42 && 37.2 & 93.32 && 45.9 & \underline{94.98} \\ 
        & AdvEnt\cite{vu2019advent}& 24.3 & 92.13 && 33.2 & 93.01 && 37.6 & 93.59 && 42.9 & 94.06 \\ 
        & s4GAN\cite{mittal2019semi}& 22.1 & 92.00 && 26.6 & 93.09 && 37.4 & 93.59 && 43.4 & 93.87 \\
        & SemiCDNet\cite{peng2021SemiCDNet} & 24.0 & \underline{92.20} && 28.3 & \cellcolor{mycyan}\textbf{93.42} && 36.2 & 92.41 && 45.3 & 94.22 \\ 
        & RCR\cite{bandaraRCR}& 27.1 & 91.63 && 32.8 & 92.99 && 36.4 & 93.07 && \underline{48.5} & 94.94 \\
        & FPA\cite{zhang2023Semisupervised}& \underline{29.0} & 91.00 && \cellcolor{mycyan}\textbf{38.2} & \underline{93.37} && \underline{39.6} & \underline{93.88} && 43.1 & 94.15 \\     
        \rowcolor{mycyan}
        \multirow{-8}{*}{\cellcolor{white}}& \cellcolor{white}AdaSemiCD   &   \textbf{30.6} & \textbf{92.52} && \cellcolor{white}\underline{33.5} & \cellcolor{white}{92.40} && \textbf{41.6} & \textbf{94.21} && \textbf{49.1} & \textbf{95.85} \\
        \cline{2-13}
        & Oracle & \multicolumn{11}{c}{$ IoU^c$=\textcolor{red}{\bf 50.1} and OA=\textcolor{red}{\bf 95.66}} \\
        \bottomrule
    \end{tabular}
    }
    \label{tab:mutil-CD}
\end{table}

\textbf{Multiclass CD Datasets:} As shown in Table \ref{tab:mutil-CD}, our proposed AdaSemiCD demonstrates optimal performance in the majority of scenarios, despite the increased complexity of multi-category change detection tasks compared to binary building change detection tasks. Notably, optimal results were achieved across all experimental settings on the CDD dataset, with $IoU^c$ and OA increasing by an average of 1.8 percentage points and 0.25 percentage points respectively, which can be attributed to its exceptionally accurate annotations, as illustrated in Fig. \ref{dataset-samples}. This high-quality labeling is particularly advantageous for pseudo-label-based semi-supervised methods. Furthermore, AdaSemiCD exhibits overall performance improvements on the SYSU-CD and CL-CD datasets. However, its performance is slightly below the SOTA results in certain experimental configurations. This can be attributed to the inherent challenges in identifying some mountain and land changes present in these datasets. These changes are often large-scale, and misclassification in such regions may result in fluctuations in performance metrics.

An additional significant observation is that, when the overall accuracy declines, the correlation between OA and $IoU^c$ becomes less straightforward. 
In particular, there are instances where $IoU^c$ reaches its maximum value, while OA does not, underscoring their different focuses. OA is mainly concerned with overall accuracy, highlighting the detection of background categories, whereas $IoU^c$ prioritizes the precise detection of target classes, which is more vital for change detection tasks than identifying background categories. This difference is apparent in the DSIFN-CD dataset, where our method demonstrates strong performance on the $IoU^c$ metric.

Among the evaluated datasets, our method demonstrates the poorest performance on DSIFN-CD. While it performs well on the critical $IoU^c$ metric, its performance on $OA$ is suboptimal. This can be attributed to similarities between DSIFN-CD and GZ-CD, as both datasets suffer from coarse manual annotations and low spatial resolution, as evident in Fig. \ref{dataset-samples}. Moreover, DSIFN-CD exhibits only a slight degree of class imbalance, with the proportion of change classes reaching as high as 34.81$\%$. Consequently, the category rebalancing strategy we designed has limited effectiveness, leading to the misclassification of many background pixels. This, in turn, contributes to the observed reduction in $OA$.

\subsubsection{Qualitative Results}

\begin{figure*}[!t]
\centering
\includegraphics[width=0.85\textwidth]{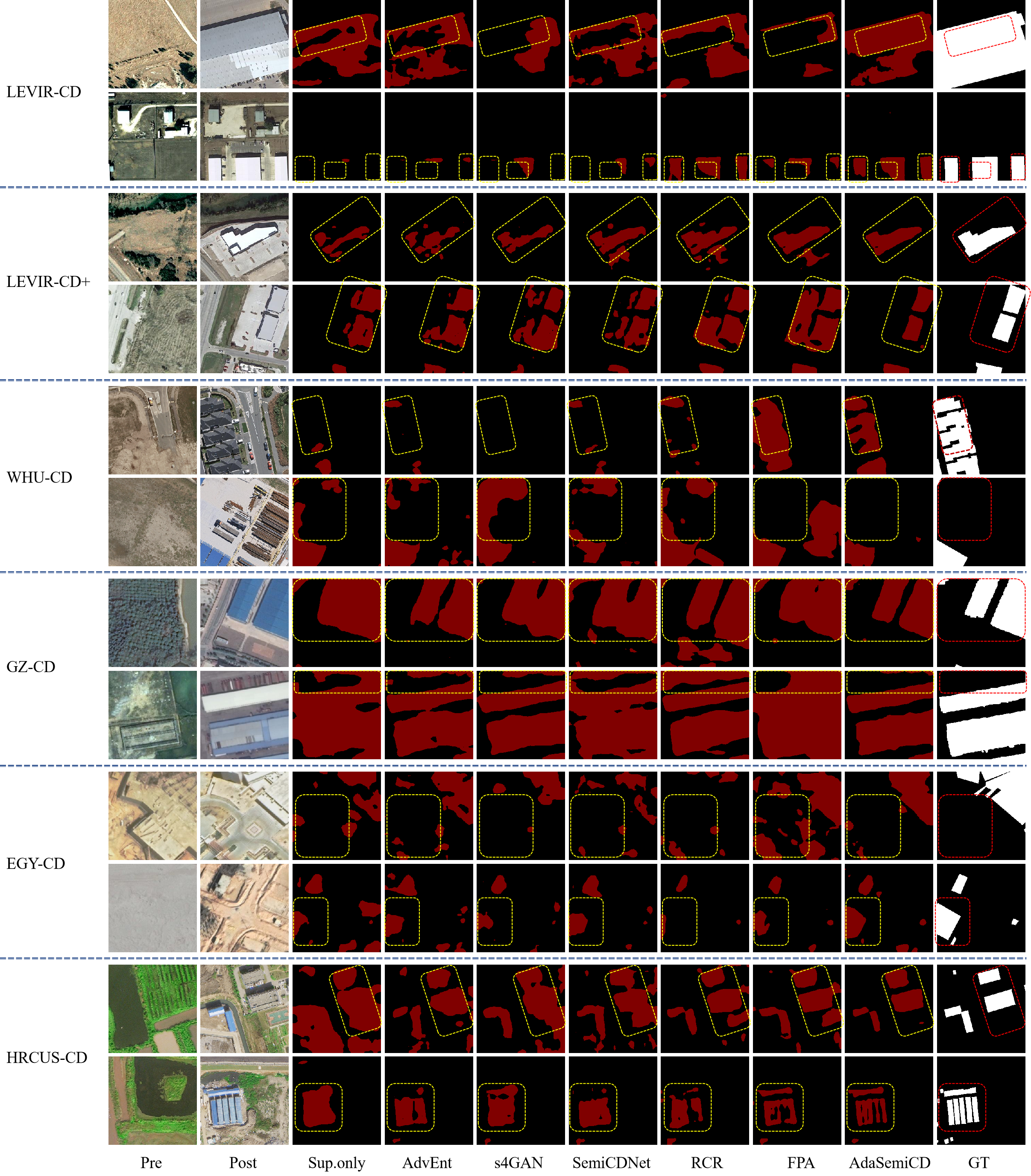}
\caption{Visualizations of different models on six building change detection datasets at the 5$\%$ labeled training ratio.}
\label{building-vis}
\end{figure*}

\begin{figure*}[!t]
\centering
\includegraphics[width=0.85\textwidth]{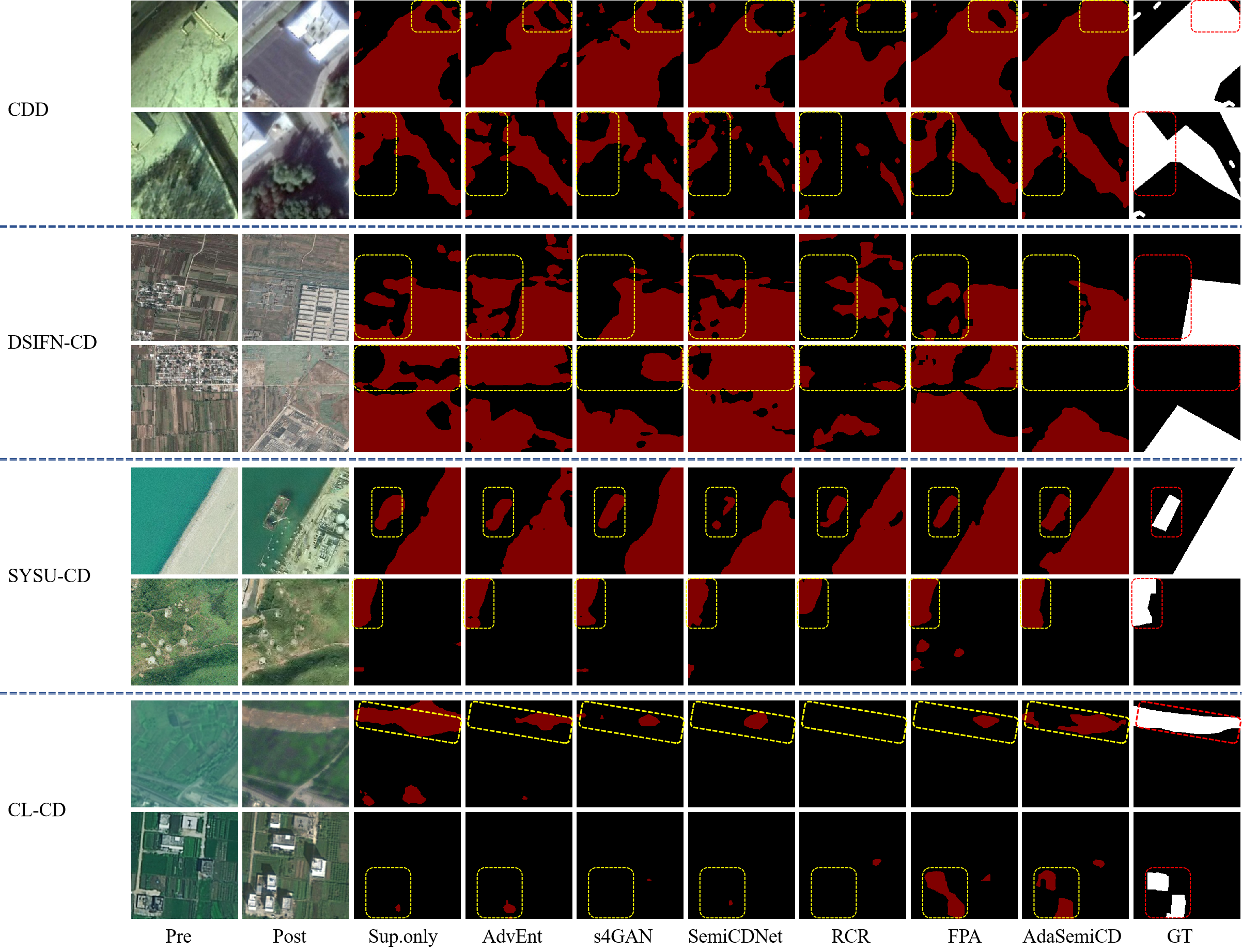}
\caption{Visualizations of different models on four mutil-class change detection datasets at the 5$\%$ labeled training ratio.}
\label{mutil-vis}
\end{figure*}

Fig. \ref{building-vis} and Fig. \ref{mutil-vis} showcase some examples of the visualizations on the test sets of building and mutilclass CD datasets respectively, in which the area selected in the box is the error-prone area. 
It is apparent that on those datasets, our approach has notably mitigated the common issues of missed and false detections. In challenging scenarios, our method could still effectively identify the areas of change that were of interest to us.

Some failure cases of the detected changes are unrelated to the buildings of interest. The absence of adequate supervision information makes it challenging to mitigate such interference, leading to decreased model performance. Additionally, the task is further complicated by the detection of small and densely changing areas, which proves to be difficult for the model.

In complex scenarios, alternative semi-supervised models may encounter guidance issues, leading them to inadvertently amplify errors during training. As a result, semi-supervised methods may perform worse than supervised methods trained exclusively on labeled data in such situations. Our model, however, adaptively mitigates these noises during the training phase by dynamically excluding them and integrating parameters from a progressively refined model throughout the training process. In these intricate cases, the quality of pseudo-labels is incrementally improved until they reach a reliable standard, thereby providing an accurate signal for model training. Consequently, our model demonstrates superior performance in these challenging regions, as highlighted by the boxes in Fig. \ref{building-vis} and Fig. \ref{mutil-vis}. This adaptive approach forms the cornerstone of our proposed method.


\subsection{Ablation Study}

\textbf{Effectiveness of proposed modules:} Due to the differences in model architecture between the current semi-supervised change detection methods referred to and ours, we did not rush to verify the superiority of our method at first. Instead, we conducted model architecture experiments first, using the classical Mean-Teacher architecture. In addition to setting hyperparameters for it to control the weight of unsupervised losses, the rest of the data augmentations and CD network remained the same as \cite{zhang2023Semisupervised} and \cite{bandaraRCR}. The gain of this semi-supervised framework compared with the single model and the two-branch network with shared weights is very obvious, and it can almost approach the previous optimal performance. This also demonstrates the validity of the principle of perturbed consistency and the parameter integration, on the basis of which we explore the adaptive training mechanism. Therefore, we separately integrated our proposed AdaEMA and AdaFusion into the MT framework and achieved average improvement of 1.2 and 5.1 on $IoU^c$ respectively, as shown in Table \ref{tab:ablation_model}. * in the table indicates that in adapt, an adaptive judgment operation is performed to determine whether the fusion is performed, and a random selection strategy is used when selecting the fusion region, and a huge gap between the two is evident. Finally, the two adaptive modules contain the complete method, and better results are obtained on the basis of individual modules, which shows that the two modules proposed by us are decoupled, and the model architecture is reasonable.

\begin{table*}[tb]
	\centering
	\caption{Ablation study of our proposed AdaSemiCD on LEVIR-CD Dataset. AEMA and AF denotes our AdaEMA, AdaFusion module respectively. And AF* represents the fusion region are randomly selected.}
     \resizebox{0.9\textwidth}{!}{
	\begin{tabular}{p{21mm}p{14mm}p{16mm}cp{14mm}p{16mm}cp{14mm}p{16mm}cp{14mm}p{16mm}} %
		\toprule
		\multirow{2}{*}{\parbox[c]{.2\linewidth}{Method}} & \multicolumn{2}{c}{5\%} & & \multicolumn{2}{c}{10\%} & & \multicolumn{2}{c}{20\%} & & \multicolumn{2}{c}{40\%}\\ 
		\cmidrule{2-3} \cmidrule{5-6} \cmidrule{8-9} \cmidrule{11-12}
		& {$IoU^c$} & {OA} && {$IoU^c$} & {OA} & & {$IoU^c$} & {OA} &&{$IoU^c$} & {OA}\\
		\midrule
		Sup. only   &   61.0 & 97.60 && 
		                66.8 & 98.13 && 
		                72.3 & 98.44 && 
		                74.9 & 98.60 \\ 
	    MT-EMA      &   67.1 {\color{red} (+6.1)} & 98.14 {\color{red}(+0.54)} && 
		                75.0 {\color{red} (+8.2)} & 98.63 {\color{red} (+0.50)} && 
		                76.6 {\color{red} (+4.3)} & 98.71 {\color{red} (+0.27)}&&
		                77.0 {\color{red} (+2.1)} & 98.73 {\color{red} (+0.13)} \\
        MT-AEMA     &   68.9 {\color{red} (+7.9)}& 98.23 {\color{red} (+0.63)} && 
		                76.1 {\color{red} (+9.3)} & 98.66 {\color{red} (+0.53)} && 
                        77.7 {\color{red} (+5.4)} & 98.78 {\color{red} (+0.34)} && 
		                77.8 {\color{red} (+2.9)} & 98.78 {\color{red} (+0.18)} \\
        (MT-EMA)+AF* &  72.0 {\color{red} (+11.0)} & 98.43 {\color{red} (+0.83)} && 
		                76.8 {\color{red} (+10.0)} & 98.72 {\color{red} (+0.59)} && 
		              77.5 {\color{red} (+5.2)} & 98.74 {\color{red} (+0.30)} && 
		                78.5 {\color{red} (+3.6)} & 98.80 {\color{red} (+0.20)} \\ 
		(MT-EMA)+AF &   77.0 {\color{red} (+16.0)} & 98.72 {\color{red} (+1.12)} && 
		                78.8 {\color{red} (+12.0)} & 98.83 {\color{red} (+0.70)} && 
		              80.4 {\color{red} (+8.1)} & 98.91 {\color{red} (+0.47)} && 
		                80.0 {\color{red} (+5.1)} & 98.90 {\color{red} (+0.30)} \\ 
        AdaSemiCD   &   77.7 {\color{red} (+16.7)} & 98.78 {\color{red} (+1.18)} && 
		                79.4 {\color{red} (+12.6)} & 98.87 {\color{red} (+0.74)} && 
		                80.3 {\color{red} (+8.0)} & 98.92 {\color{red} (+0.48)} &&
		                80.6 {\color{red} (+5.7)} & 98.93 {\color{red} (+0.33)} \\     
		\bottomrule
	\end{tabular}
    }
	\label{tab:ablation_model}
\end{table*}

\textbf{Pseudo-label qualification metric}: As described in Section \ref{pseudo-label-metric}, the pseudo-label evaluation metric we proposed is based on information entropy, enhanced by class rebalancing and confusion region amplification. To demonstrate the effectiveness, we first evaluated our AdaSemiCD using information entropy as the sole metric, then incrementally incorporated class rebalancing and confusion region amplification, and finally evaluated the full model, which integrates all components.

The experimental results are presented in Table \ref{tab:ablation_metric}. When only information entropy is used as the evaluation metric for implementing the adaptive training mechanism, significant performance improvements are achieved, attributed to the AdaFusion and AdaEMA modules we designed. Incorporating class rebalancing and confusion region amplification based on information entropy further enhances performance, demonstrating that these improvements enable more accurate identification of pseudo-labels during training, thereby facilitating more precise adaptive operations. Notably, the gain from class rebalancing exceeds that from confusion region amplification. This is because, with class rebalancing, the evaluation of pseudo labels places greater emphasis on the foreground, somewhat neglecting background identification, thus allowing AdaFusion to focus more on the uncertainty region in the foreground. In addition, confusion region amplification proves more beneficial as more labeled data becomes available. Finally, using uncertainty, an overall assessment that includes both components, leads to further performance gains, suggesting that these two improvements are independent and can be effectively combined for more accurate assessments.
\begin{table*}[tb]
	\centering
	\caption{Ablation study of pseudo-label qualification metric on LEVIR-CD Dataset.}
     \resizebox{0.9\textwidth}{!}{
	\begin{tabular}{p{21mm}p{14mm}p{16mm}cp{14mm}p{16mm}cp{14mm}p{16mm}cp{14mm}p{16mm}} %
		\toprule
		\multirow{2}{*}{\parbox[c]{.2\linewidth}{Method}} & \multicolumn{2}{c}{5\%} & & \multicolumn{2}{c}{10\%} & & \multicolumn{2}{c}{20\%} & & \multicolumn{2}{c}{40\%}\\ 
		\cmidrule{2-3} \cmidrule{5-6} \cmidrule{8-9} \cmidrule{11-12}
		& {$IoU^c$} & {OA} && {$IoU^c$} & {OA} & & {$IoU^c$} & {OA} &&{$IoU^c$} & {OA}\\
		\midrule
		Sup. only   &   61.0 & 97.60 && 
		                66.8 & 98.13 && 
		                72.3 & 98.44 && 
		                74.9 & 98.60 \\ 
	   Entropy      &   73.2 {\color{red} (+12.2)} & 98.50 {\color{red}(+0.90)} && 
		                77.4 {\color{red} (+10.6)} & 98.75 {\color{red} (+0.62)} && 
		                78.6 {\color{red} (+6.3)} & 98.81 {\color{red} (+0.37)}&&
		                79.2 {\color{red} (+4.3)} & 98.89 {\color{red} (+0.29)} \\
        Entropy+rebalance     &   76.0 {\color{red} (+15.0)}& 98.63 {\color{red} (+1.03)} && 
		                78.7 {\color{red} (+11.9)} & 98.82 {\color{red} (+0.69)} && 
                        79.6 {\color{red} (+7.3)} & 98.85 {\color{red} (+0.41)} && 
		                79.6 {\color{red} (+4.7)} & 98.86 {\color{red} (+0.26)} \\
        Entropy+confusion &  74.6 {\color{red} (+13.6)} & 98.61 {\color{red} (+1.01)} && 
		                78.0 {\color{red} (+11.2)} & 98.79 {\color{red} (+0.66)} && 
		              79.2 {\color{red} (+6.9)} & 98.82 {\color{red} (+0.38)} && 
		                79.8 {\color{red} (+4.9)} & 98.87 {\color{red} (+0.27)} \\ 
        Uncertainty   &   77.7 {\color{red} (+16.7)} & 98.78 {\color{red} (+1.18)} && 
		                79.4 {\color{red} (+12.6)} & 98.87 {\color{red} (+0.74)} && 
		                80.3 {\color{red} (+8.0)} & 98.92 {\color{red} (+0.48)} &&
		                80.6 {\color{red} (+5.7)} & 98.93 {\color{red} (+0.33)} \\     
		\bottomrule
	\end{tabular}
    }
	\label{tab:ablation_metric}
\end{table*}

\begin{figure}[!t]
\includegraphics[width=0.85\columnwidth]{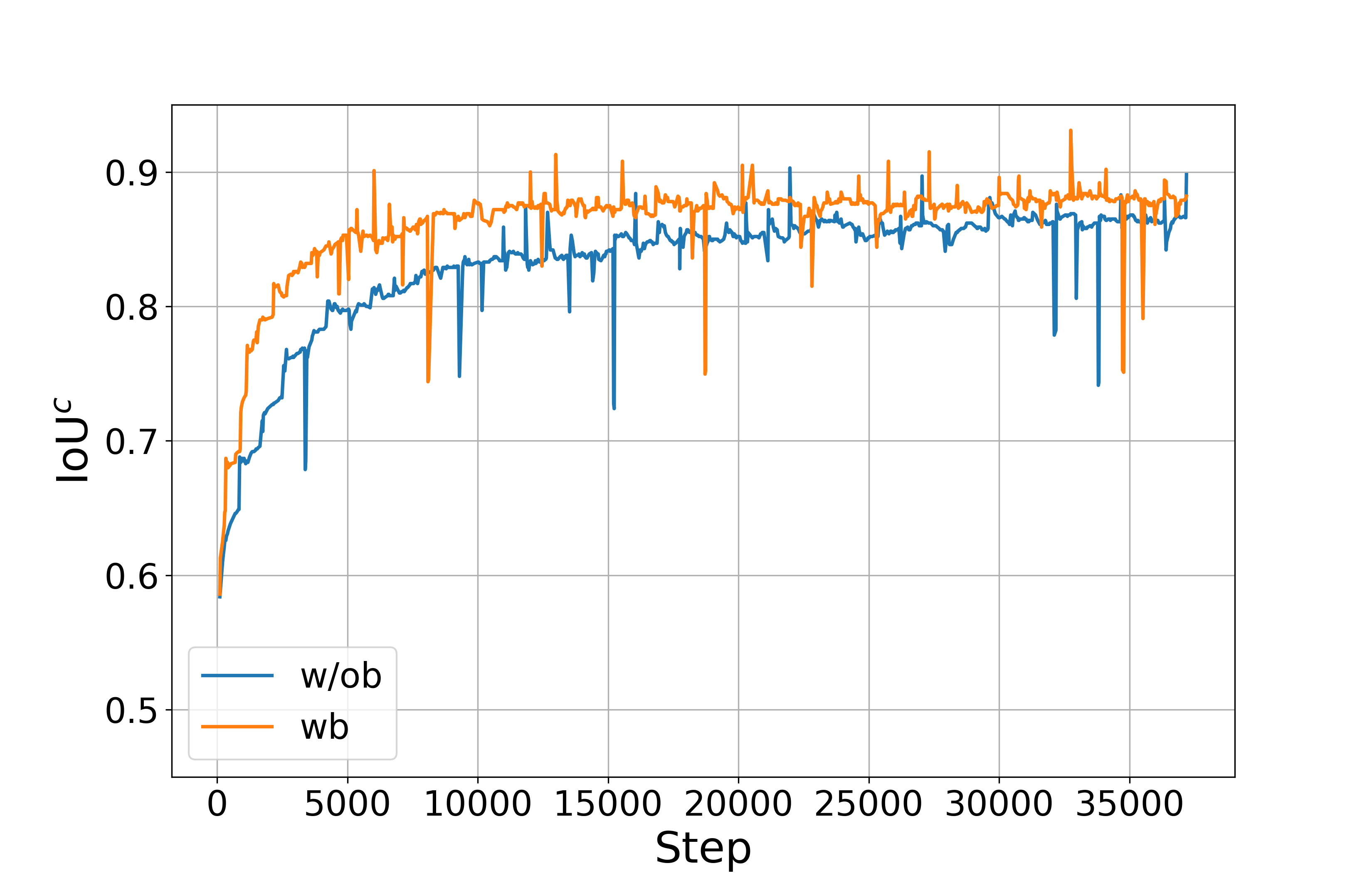}
\caption{The influence of class rebalancing on the quality of pseudo-labels generated for unlabeled samples during training.}
\label{Un_iou}
\end{figure}

We have stored the pseudo-labels generated during the training process, and Fig.~\ref{Un_iou} illustrates the $IoU^c$ between the pseudo-labels and the corresponding Ground Truth throughout training, both with and without class rebalancing in the pseudo-label evaluation (note: the Ground Truth used here for unlabeled samples are solely for calculating the $IoU^c$ with the pseudo-labels and are not involved in any other aspect of the training process). It is evident that, although class rebalancing does not directly improve the pseudo-labels, it enables a more accurate evaluation of the pseudo-labels. As a result, our adaptive training mechanism enhances the quality of them, bringing them closer to the real labels, which leads to a higher $IoU^c$ between them.

\begin{table}[tb]
	\centering
	\caption{Sensitivity analysis of ramp-up hyperparameters with 10$\%$ labeled data on the LEVIR-CD, WHU-CD, and CDD datasets.}
     \resizebox{0.9\columnwidth}{!}{
	\begin{tabular}{p{3mm}p{10mm}p{5mm}cp{1mm}p{5mm}cp{1mm}p{5mm}cp{1mm}p{5mm}} %
		\toprule
		\multirow{2}{*}{\parbox[c]{.1\linewidth}{$\gamma$}} & \multirow{2}{*}{$w_{max}$} & \multicolumn{2}{c}{LEVIR-CD} & & \multicolumn{2}{c}{WHU-CD} & & \multicolumn{2}{c}{CDD}\\ 
	  \cmidrule{3-4} \cmidrule{6-7} \cmidrule{9-10}
		&& {$IoU^c$} & {OA} && {$IoU^c$} & {OA} & & {$IoU^c$} & {OA} \\
		\midrule
  	0  & 0 &   66.8 & 98.13 && 
		                55.7 & 97.53 && 
		                67.9 & 95.46 \\  
		0.05  & 1.0 &   67.2 & 98.17 && 
		                53.8 & 97.02 && 
		                74.4 & 96.35 \\  
	    0.1   & 1.0   &   71.8 & 98.32 && 
		                61.0 & 98.10 && 
		                \underline{\textbf{77.3}} & \underline{\textbf{96.89}} \\  
        0.3   & 1.0   &   69.9 & 98.26 && 
		                59.4 & 98.03 && 
		                76.2 & 96.56 \\  
        0.5   & 1.0   &  68.7 & 98.15 && 
		              60.9 & 98.25 && 
		                76.3 & 96.58 \\  
		1.0   & 1.0   &   67.3 & 98.13 && 
		                60.5 & 98.18 && 
		                72.3 & 95.98 \\ 
        0.1   & 0.1   &   65.2 & 97.60 && 
		                \underline{\textbf{70.8}} & \underline{\textbf{98.70}} && 
		                71.6 & 95.77 \\  
        0.1   & 0.5   &   68.3 & 98.14 && 
		                66.9 & 98.54 && 
		                75.8 & 96.67 \\  
        0.1   & 5.0   &   73.9 & 98.75 && 
		                60.1 & 98.00 && 
		                69.1 & 95.51 \\  
        0.1   & 10.0   & \underline{\textbf{79.4}} & \underline{\textbf{98.87}} && 
		                52.4 & 97.40 && 
		                68.2 & 95.50 \\  
        0.1   & 30.0   &   71.9 & 98.42 && 
		                50.34 & 97.12 && 
		                65.4 & 95.20 \\  
		\bottomrule
	\end{tabular}
    }
	\label{tab:ablation_gamma}
\end{table}

\subsection{Complexity Analysis}
Since we utilize the same architecture with FPA and RCR,  the number of training parameters (46.85M) and computational amount (585.85 GFLOPs) were the same. The variations in parameters, FLOPs, and the time required for training and inference are presented in Table \ref{tab:compute}. Our AdaSemiCD is comparable to other methods with reference count, FLOPs, and inference time. The main reason for the longer training time was that it took about 0.3s for each iteration to generate and evaluate pseudo-labels twice, while it only took about 0.006s and 0.03s for fusion and EMA parameters updating, respectively. Moreover, our model balances performance and time consumption, offering a notable performance benefit with only a slight increase in time.
\begin{table}[h]
    \centering
    \caption{COMPARISON OF PARAMETERS, COMPUTING COMPLEXITY, AND TRAINING TIME OF DIFFERENT SSCD METHODS ON 5$\%$ LABELED LEVIR-CD Dataset.}
    \label{tab:compute} 
     \resizebox{0.48\textwidth}{!}{
    \begin{tabular}{p{20mm}p{15mm}p{15mm}p{15mm}p{15mm}p{10mm}} %
    \toprule 
    \multirow{2}{*}{Method} & \multirow{2}{*}{Params(M)} & \multirow{2}{*}{FLOPs(G)} & Training Time(s) &Inference Time(ms) &\multirow{2}{*}{{$IoU^c$}}\\
    \midrule 
    Sup.Only & \hspace{0.15cm}46.85 & \hspace{0.10cm}585.85 & \hspace{0.35cm}77 & \hspace{0.4cm}56 & 61.0\\
    AdvEnt\cite{vu2019advent} & \hspace{0.15cm}46.85 & \hspace{0.10cm}585.85 & \hspace{0.35cm}405 & \hspace{0.4cm}63 & 66.1\\
    s4GAN\cite{mittal2019semi} & \hspace{0.15cm}46.85 & \hspace{0.10cm}585.85 & \hspace{0.35cm}585 & \hspace{0.4cm}58 & 64.0\\
    SemiCDNet\cite{peng2021SemiCDNet} & \hspace{0.15cm}46.85 & \hspace{0.10cm}585.85 & \hspace{0.35cm}408 & \hspace{0.4cm}75 & 67.6\\    
    RCR\cite{bandaraRCR} & \hspace{0.15cm}46.85 & \hspace{0.10cm}585.85 & \hspace{0.35cm}742 & \hspace{0.4cm}59 & 72.5\\
    FPA\cite{zhang2023Semisupervised} & \hspace{0.15cm}46.85 & \hspace{0.10cm}585.85 & \hspace{0.35cm}727 & \hspace{0.4cm}68 & 73.7\\
    AdaSemiCD & \hspace{0.15cm}46.85 & \hspace{0.10cm}585.85 & \hspace{0.35cm}915 & \hspace{0.4cm}67 &   \textbf{77.7}\\
    \hline
    Oracle & \hspace{0.15cm}46.85 & 
    \hspace{0.10cm}585.85 & \hspace{0.35cm}293 & \hspace{0.4cm}55 & 77.9 \\
    \bottomrule 
\end{tabular}
}
\end{table}
\textbf{Hyperparameters in Ramp-up:} The Ramp-up process has a significant influence on the performance of our AdaSemiCD on SSCD. Therefore, we conduct experiments on the selection of two hyperparameters ($\gamma$ and $w_{max}$) that control the Ramp-up process. As shown in Table \ref{tab:ablation_gamma}, our method achieves the best performance on the three datasets under the combination of parameters (0.1, 10), (0.1, 0.1), and (0.1, 1.0) respectively. Moreover, our method is sensitive to this hyperparameter, and inappropriate parameter selection will cause large performance attenuation. This is because our method conducts supervised training on labeled samples and unsupervised training on unlabeled samples at the same time. If the relationship between the two cannot be properly balanced, overfitting on labeled samples or excessive noise interference from unlabeled samples will be caused. All of our remaining experiments were performed at this hyperparameter setting, and the hyperparameters of the compared methods were consistent with the best choices in their original paper.

\section{Conclusion}
In this study, we present AdaSemiCD, a flexible semi-supervised framework for change detection. This framework assesses the quality of pseudo-labels on unlabeled training samples and implements adaptive modifications based on the assessment outcomes, which include sample fusion (AdaFusion), and parameter updates (AdaEMA). Despite the complexity of the scenes, our model successfully identifies the areas of interest with minimal interference during training. Empirical evidence from ten publicly available CD datasets attests to the efficacy of our methodology. Looking ahead, this adaptive processing technique holds promise for potential application in other semi-supervised tasks.

\bibliography{ref.bib}

\vfill

\begin{IEEEbiography}
[{\includegraphics[width=1in,height=1.25in,clip,keepaspectratio]{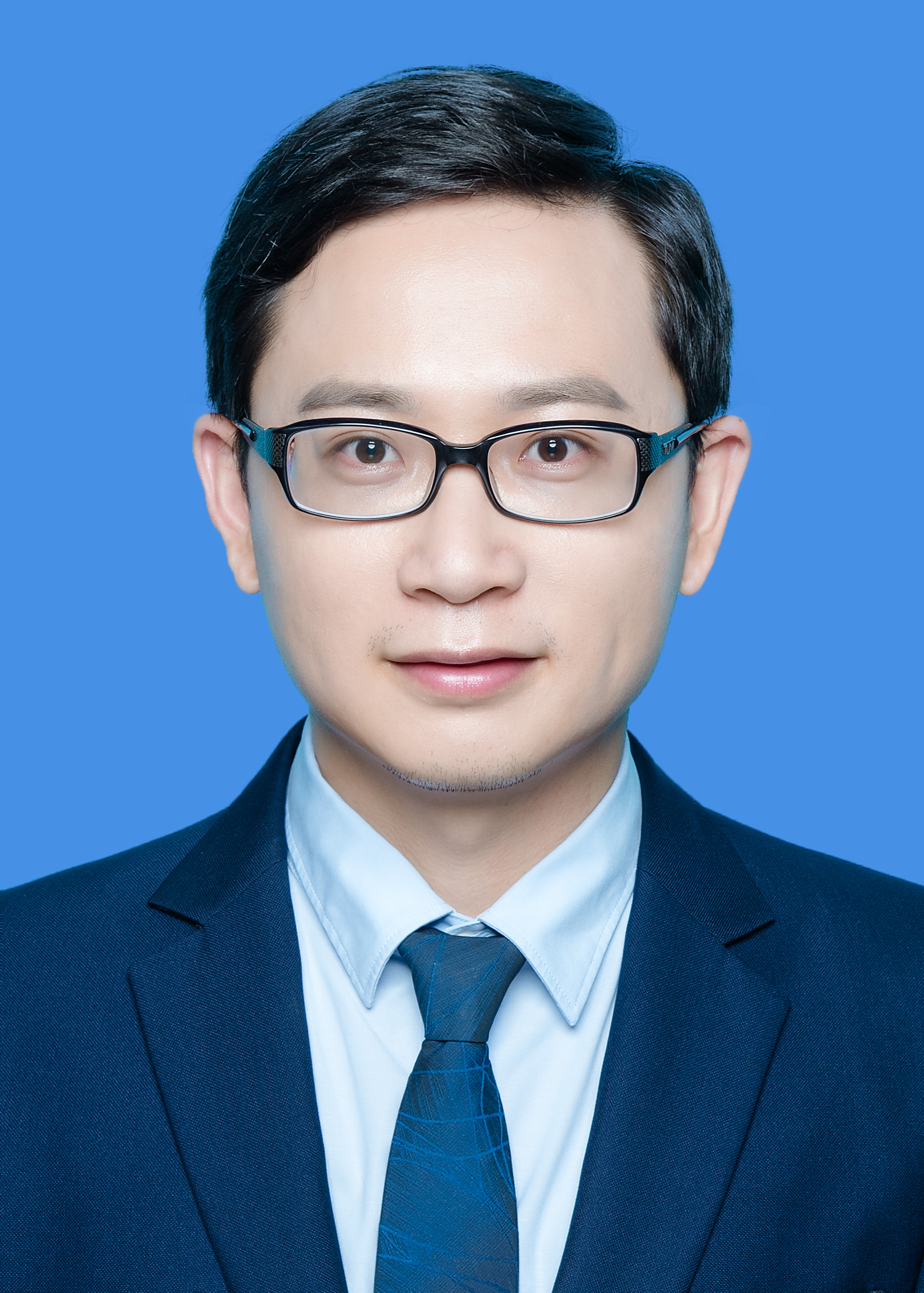}}]
{Lingyan Ran} received his B.S. and Ph.D. degrees from Northwestern Polytechnical University (NWPU), Xi'an, China, in 2011 and 2018. Earlier, he was a visiting scholar at Stevens Institute of Technology, Hoboken, NJ, from 2013 to 2015. He is currently an Associate Professor with the School of Computer Science, NWPU. His research interests include image classification, semantic segmentation, and change detection. 
\end{IEEEbiography}
\begin{IEEEbiography}
[{\includegraphics[width=1in,height=1.25in,clip,keepaspectratio]{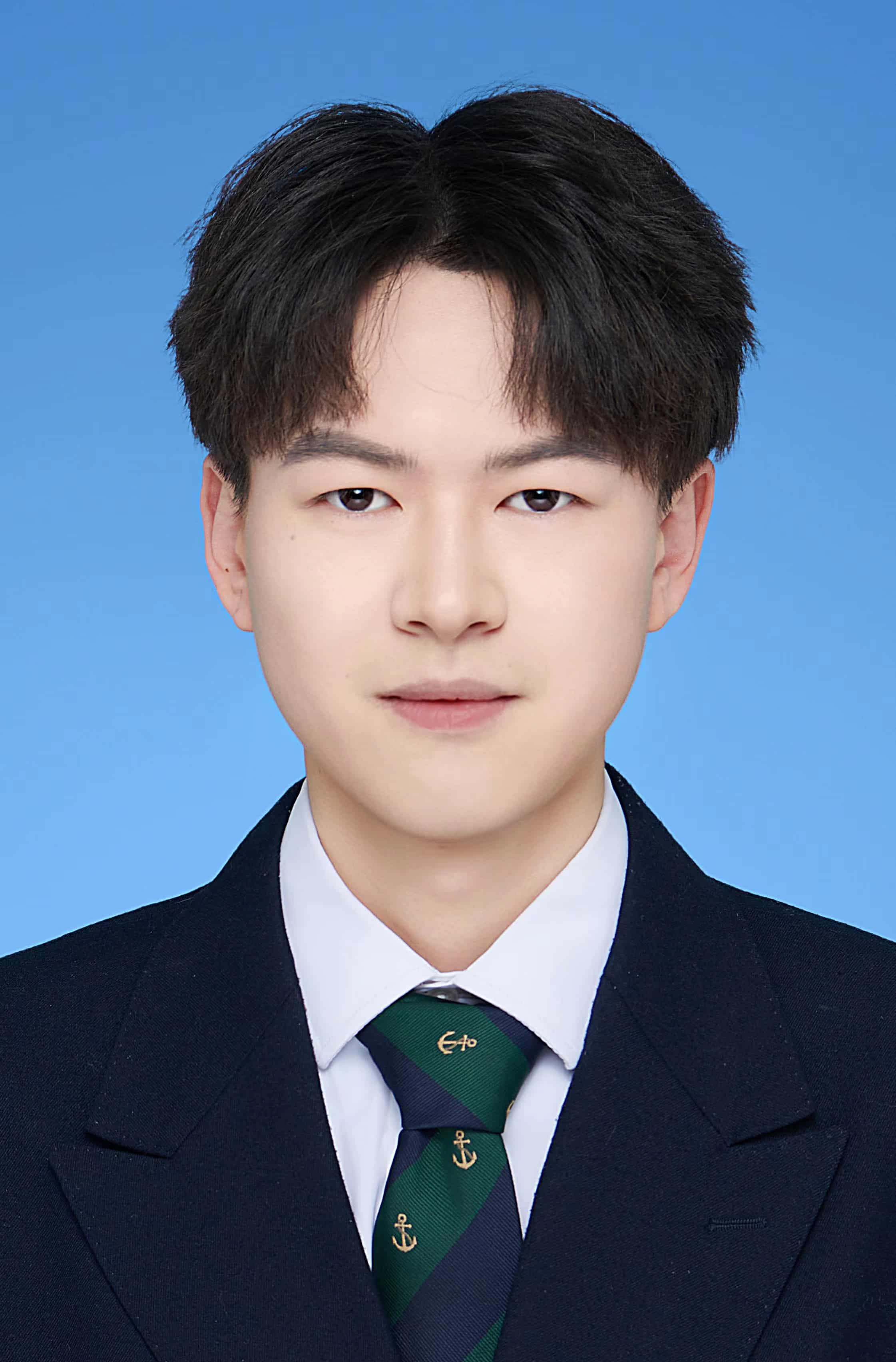}}]
{Dongcheng Wen} received the B.S. degree in Beijing Technology and Business University, Beijing, China, in 2022. 

He is currently pursuing the master’s degree with the National Engineering Laboratory for Integrated Aero-Space-Ground-Ocean Big Data Application Technology, Northwestern Polytechnical University, Xi’an, China. His research interests include the change detection of remote sensing images. 
\end{IEEEbiography}
\begin{IEEEbiography}
[{\includegraphics[width=1in,height=1.25in,clip,keepaspectratio]{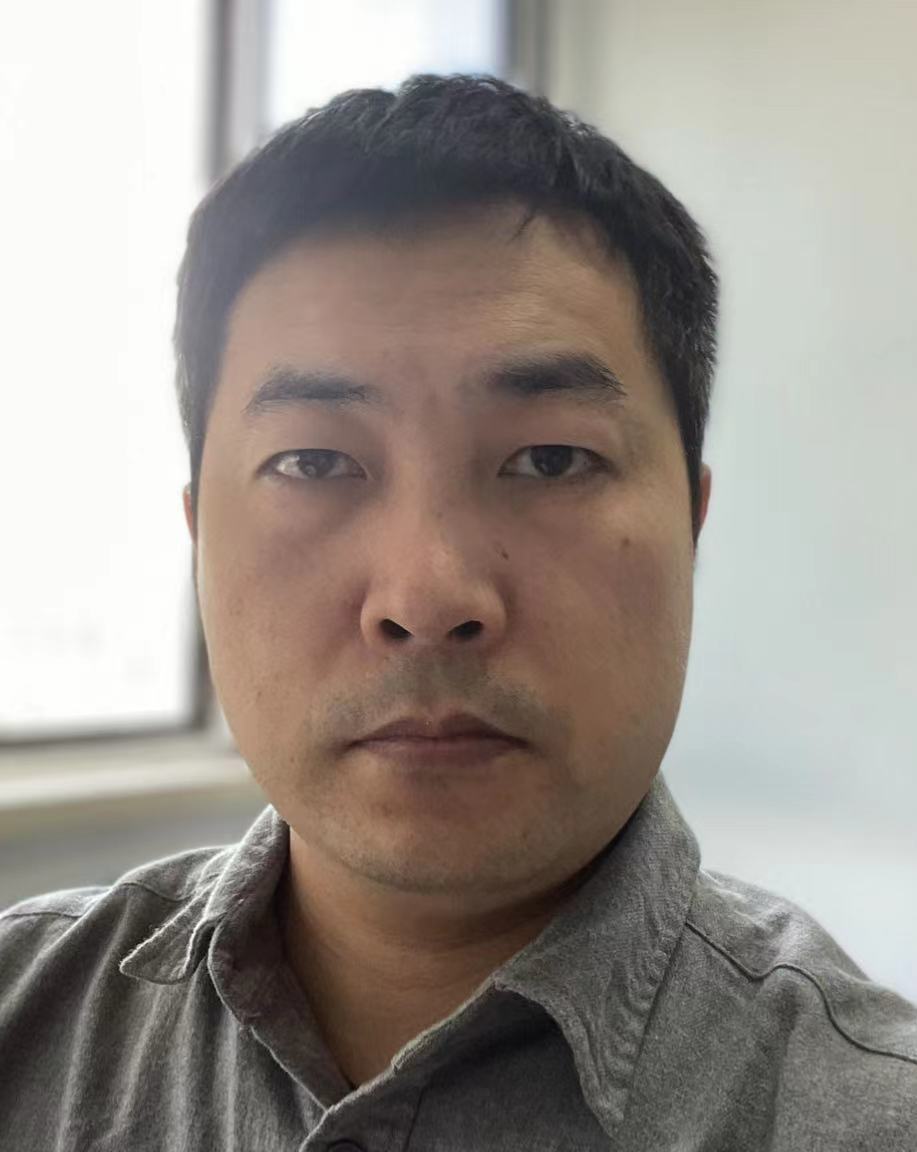}}]
{Tao Zhuo} received the Ph.D. degree in computer science and technology from Northwestern Polytechnical University, Xi‘an, China, in 2016. From 2016 to 2021, he was a Research Fellow with the National University of Singapore, Singapore. From 2021 to 2024, he was with Shandong Artificial Intelligence Institute, Jinan, China. 

He is currently a Professor with the College of Information Engineering, Northwest A\&F University, Xianyang, China. His research interests include image/video processing, computer vision, and machine learning.
\end{IEEEbiography}
\begin{IEEEbiography}
[{\includegraphics[width=1in,height=1.25in,clip,keepaspectratio]{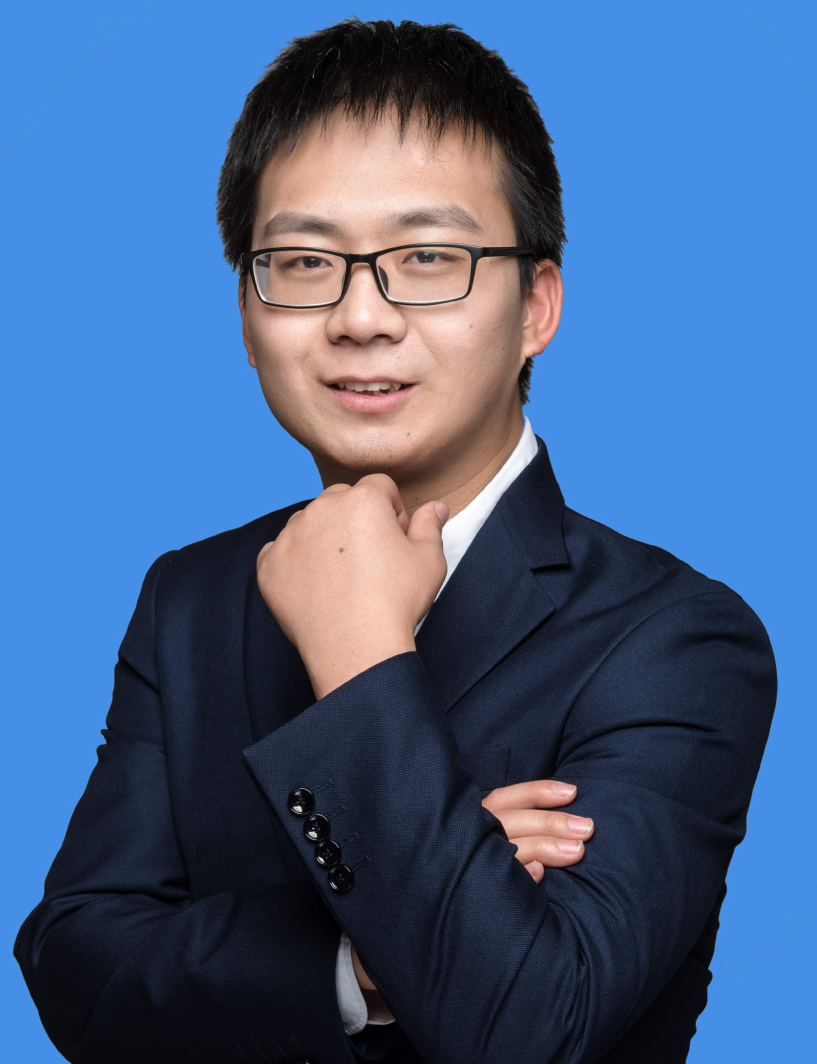}}]
{Shizhou Zhang} received the B.E. and Ph.D. degrees from Xi’an Jiaotong University, Xi’an, China, in 2010, and 2017, respectively. 

He is currently with Northwestern Polytechnical University, Xi’an, as an Associate Professor (Tenured). His research interests include content based image analysis, pattern recognition, and machine learning, specifically in the areas of deep learning-based vision tasks, such as image classification, object detection, reidentification, and neural architecture search. 
\end{IEEEbiography}
\begin{IEEEbiography}
[{\includegraphics[width=1in,height=1.25in,clip,keepaspectratio]{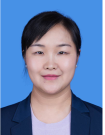}}]
{Xiuwei Zhang} received the B.S., M.S., and Ph.D. degrees from the School of Computer Science, Northwestern Polytechnical University, Xi’an, China, in 2004, 2007, and 2011, respectively. 

She is currently an Associate Professor with the School of Computer Science, Northwestern Polytechnical University. Her research interests include remote sensing image processing, multimodel image fusion, image registration, and intelligent forecasting. 
\end{IEEEbiography}
\begin{IEEEbiography}
[{\includegraphics[width=1in,height=1.25in,clip,keepaspectratio]{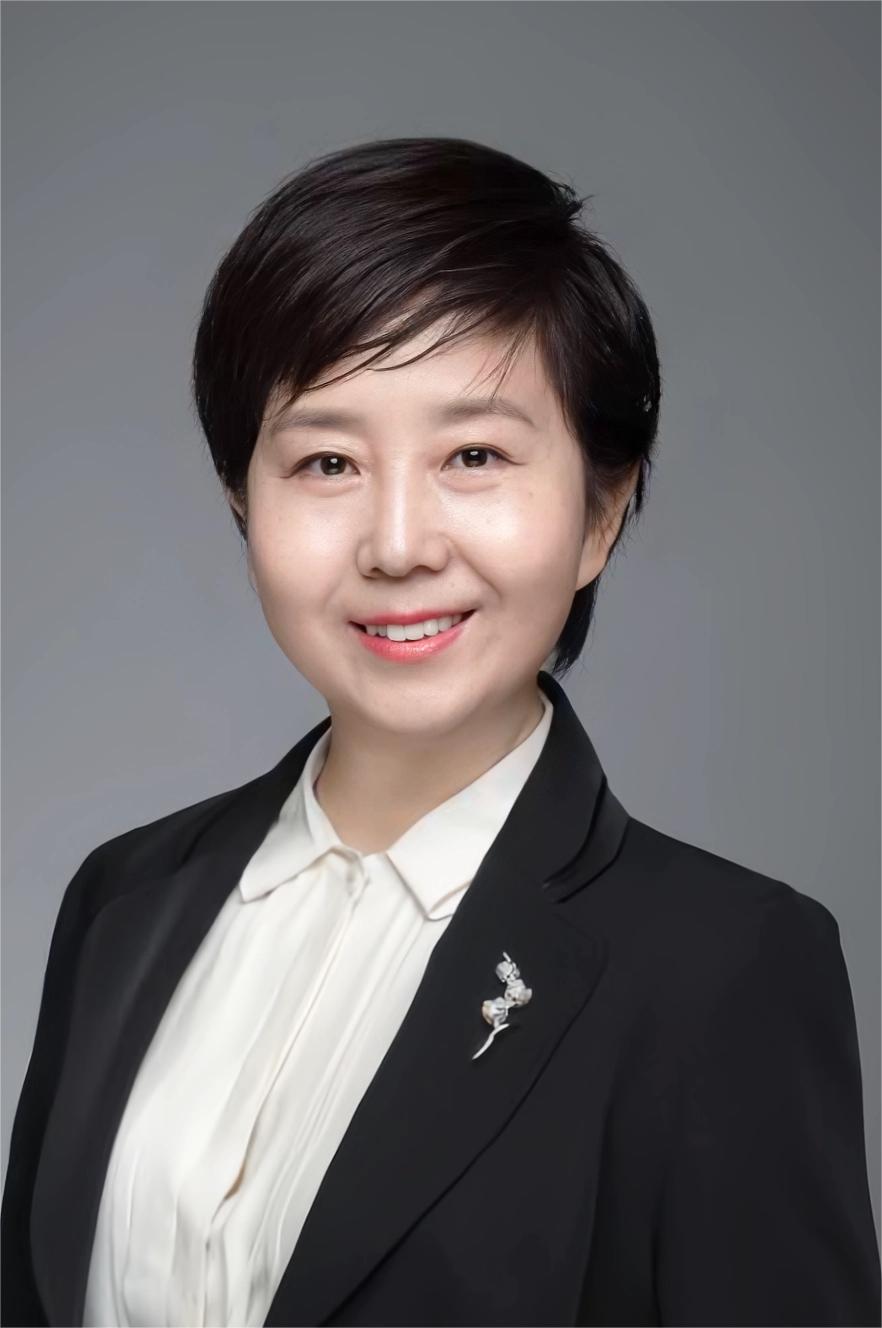}}]
{Yanning Zhang} (Fellow, IEEE) received the B.S. degree from Dalian University of Science and Engineering in 1988, and the M.S. and Ph.D. degrees from Northwestern Polytechnical University, Xi’an, China, in 1993 and 1996, respectively. 

She is currently a Professor with the School of Computer Science, Northwestern Polytechnical University. She has published over 200 articles in international journals, conferences, and Chinese key journals. Her research interests include signal and image processing, computer vision, and pattern recognition. 
\end{IEEEbiography}

\end{document}